\documentclass[sigconf]{acmart}

\copyrightyear{2024}
\acmYear{2024}
\setcopyright{rightsretained}
\acmConference[CIKM '24]{Proceedings of the 33rd ACM International Conference on Information and Knowledge Management}{October 21--25, 2024}{Boise, ID, USA}
\acmBooktitle{Proceedings of the 33rd ACM International Conference on Information and Knowledge Management (CIKM '24), October 21--25, 2024, Boise, ID, USA}
\acmDOI{10.1145/3627673.3679686}
\acmISBN{979-8-4007-0436-9/24/10}

\newcommand{\proposed}{\textsf{Metacon}}

\usepackage{microtype,textcase}
\usepackage{mathtools,xparse}
\usepackage{amsthm, physics}
\usepackage{amsmath}
\usepackage{algorithm,enumitem}
\usepackage{algpseudocode}
\usepackage{algorithmicx}
\usepackage{dsfont}

\usepackage{multirow}
\usepackage{graphicx}
\usepackage{booktabs}
\usepackage{pifont}
\usepackage{xcolor}
\usepackage{blindtext}
\usepackage{cleveref}
\definecolor{airforceblue}{rgb}{0.36, 0.54, 0.66}
\definecolor{burntumber}{rgb}{0.54, 0.2, 0.14}

\begin{document}
\title{Debiased Graph Poisoning Attack via Contrastive Surrogate Objective}
  
\author{Kanghoon Yoon}
\affiliation{%
    \institution{KAIST}
  \city{Daejeon}
  \country{Republic of Korea}}
\email{ykhoon08@kaist.ac.kr}

\author{Yeonjun In}
\affiliation{%
    \institution{KAIST}
  \city{Daejeon}
  \country{Republic of Korea}}
\email{yeonjun.in@kaist.ac.kr}

\author{Namkyeong Lee}
\affiliation{%
    \institution{KAIST}
  \city{Daejeon}
  \country{Republic of Korea}}
\email{namkyeong96@kaist.ac.kr}

\author{Kibum Kim}
\affiliation{%
    \institution{KAIST}
  \city{Daejeon}
  \country{Republic of Korea}}
\email{kb.kim@kaist.ac.kr}

\author{Chanyoung Park}
\authornote{Corresponding author.}
\affiliation{%
    \institution{KAIST}
  \city{Daejeon}
  \country{Republic of Korea}}
\email{cy.park@kaist.ac.kr}

\acmArticleType{Research}
\acmCodeLink{https://github.com/KanghoonYoon/torch-metacon}

\begin{abstract}
\looseness=-1
Graph neural networks (GNN) are vulnerable to adversarial attacks, which aim to degrade the performance of GNNs through imperceptible changes on the graph. 
However, we find that in fact the prevalent meta-gradient-based attacks, which utilizes the gradient of the loss w.r.t the adjacency matrix, are biased towards training nodes. That is, their meta-gradient is determined by a training procedure of the surrogate model, which is solely trained on the training nodes. This bias manifests as an uneven perturbation, connecting two nodes when at least one of them is a labeled node, i.e., training node, while it is unlikely to connect two unlabeled nodes. However, these biased attack approaches are sub-optimal as they do not consider flipping edges between two unlabeled nodes at all. This means that they miss the potential attacked edges between unlabeled nodes that significantly alter the representation of a node.
In this paper, we investigate the meta-gradients to uncover the root cause of the uneven perturbations of existing attacks. 
Based on our analysis, we propose a \textbf{\textsf{Meta}}-gradient-based attack method using \textbf{\textsf{con}}trastive surrogate objective (\proposed), which alleviates the bias in meta-gradient using a new surrogate loss.
We conduct extensive experiments to show that~\proposed~outperforms existing meta gradient-based attack methods through benchmark datasets, while showing that alleviating the bias towards training nodes is effective in attacking the graph structure. The code for~\proposed~is available at \url{https://github.com/KanghoonYoon/torch-metacon}.
\end{abstract}



\maketitle

\section{Introduction}
\label{section:intro}

Graph Neural Network (GNN) is a powerful representation learner that captures both patterns of node features and structured information. Benefiting from advances in GNNs, several applications such as social network analysis, prediction of molecular properties, and recommendation systems are seeing noteworthy growth~\cite{iclr2017kipf-gcn,lee2023shiftmol,Wang_2019gcf,hawkes}. Recently, seminal works demonstrate that GNNs are vulnerable to adversarial attacks, which perturb node features or the graph structure to disrupt the prediction of victim GNN models \cite{pmlr-rl2sv-advattack-on-graph, ijcai19-advexamples-for-graph, kdd22-spectral-attack}. To address the vulnerability of GNNs to adversarial attacks, there have been studies on defense methods that learn graph representations that are robust against small changes in the graph.
\cite{kdd19-rgcn, kdd20-prognn, neurips20_gnnguard, neurips20_ReliableAGG, wsdm22-simpgcn, wsdm22-rsgnn, in2023spagcl, in2024sggsr}. While progress has been made in defending against adversarial attacks on graphs, little attention has been paid to understanding the graph attack methods per se.

Graph adversarial attacks aim to degrade the performance of GNNs through making imperceptible changes to the graph. Specifically, these attacks alter the node features or the graph structure~\cite{iclr18-mettack}, inject nodes~\cite{www20-node-injection-nipa}, and rewire edges~\cite{kdd21-rewire-attack}. Among various strategies for attacking graphs, our interest lies in graph structure attacks, and particularly \textit{untargeted attack methods} \cite{kdd21-adv-attack-survey}, whose goal is to drop the overall performance on testing nodes, as they are widely taken as a baseline attack method against which recent robust GNN methods attempt to defend \cite{kdd19-rgcn, kdd20-prognn, neurips20_gnnguard, wsdm22-rsgnn}. 

\looseness=-1
Recent advancements in graph adversarial attack methods have demonstrated significant effectiveness in deteriorating the generalization capability of GNNs, which is largely attributed to the use of \emph{meta-gradients}. 
The core idea behind the meta-gradient-based attack is to manipulate edges based on the gradients of the loss with respect to the adjacency matrix. More specifically, based on the gradient information, the attacker sequentially flips the edges that greatly increase the loss. Using this principle, MetaAttack~\cite{iclr18-mettack} first introduces the meta-gradient to the graph structure, which is obtained by backpropagating through the learning procedure of a differentiable model, e.g., GNN. In particular, MetaAttack provokes a potent performance drop of the victim models by alternatively training a surrogate model with the attacked graph, and obtaining the meta-gradients based on the trained surrogate model. Encouraged by the effectiveness of the meta-gradient, EpoAtk \cite{icdm20-epoatk} designs an exploratory attack method with a greedy algorithm to expand the attack search space of meta-gradients, while AtkSE \cite{cikm22-atkse} stabilizes the uncertainty of the meta-gradient to establish a consistent meta-gradient-based attack method. 
Moreover, GraD \cite{liu2022grad_towards} introduces a new attack loss that encourages the attack methods to flip edges near correctly classified nodes, which in turn deteriorates the victim GNN models by causing confusion among a large number of correctly classified nodes. 

\begin{table}[t]
    \centering
    \captionof{table}{(a) Number of attacks (i.e., flipped edges) and (b) the ratio of attacks in each edge set
    (i.e., $\text{ratio}=\frac{\text{\# of attacks}}{\text{\# clean edges in an edge set}}\times 100$), where $5\%$ edges are perturbed on Citeseer dataset, and training portion is $10\%$. Meta-gradient-based attacks unevenly perturb the graph structure, while PGD-based attacks do not.}
    \vspace{-2.2ex}
    \label{tab:intro}
    \scalebox{0.72}{
    \begin{tabular}{c||c|ccc}
    \hline
     & Model & L-L & L-U & U-U  \\
    \hline
    \multirow{6}{.21\linewidth}{(a) Num. Attacks} & MetaAttack & 4.5 & 177.5 & 0.5 \\ 
    
    & EpoAtk & 31.5 & 151.0 & 0.0 \\ 
    
    & AtkSE & 20.0  & 162.5  & 0.0  \\ 
    
    & GraD & 6.0  & 175.0  & 0.5  \\ \cline{2-5}

    & PGD-CE & 4.0  & 54.0  & 121.5 \\ 
    & PGD-CW & 7.0  & 55.0  & 115.5 \\ \hline \hline

    \multirow{6}{.21\linewidth}{(b) Attack Ratio (\%)} & MetaAttack & 15.1 & 29.1 & \textbf{0.0} \\ 

    & EpoAtk & 93.9 & 24.8 & \textbf{0.0} \\ 

    & AtkSE & 64.0 & 26.6 & \textbf{0.0} \\ 

    & GraD  & 10.2 & 28.7 & \textbf{0.0} \\ \cline{2-5}

    & PGD-CE & 13.1 & 8.9  & 4.0 \\ 
    & PGD-CW & 23.9  & 9.1  & 3.8 \\ \hline
        
    \end{tabular}
    }
    \vspace{-3ex}
\end{table}

\begin{figure}[t]
    \centering
    \includegraphics[width=1.0\columnwidth]{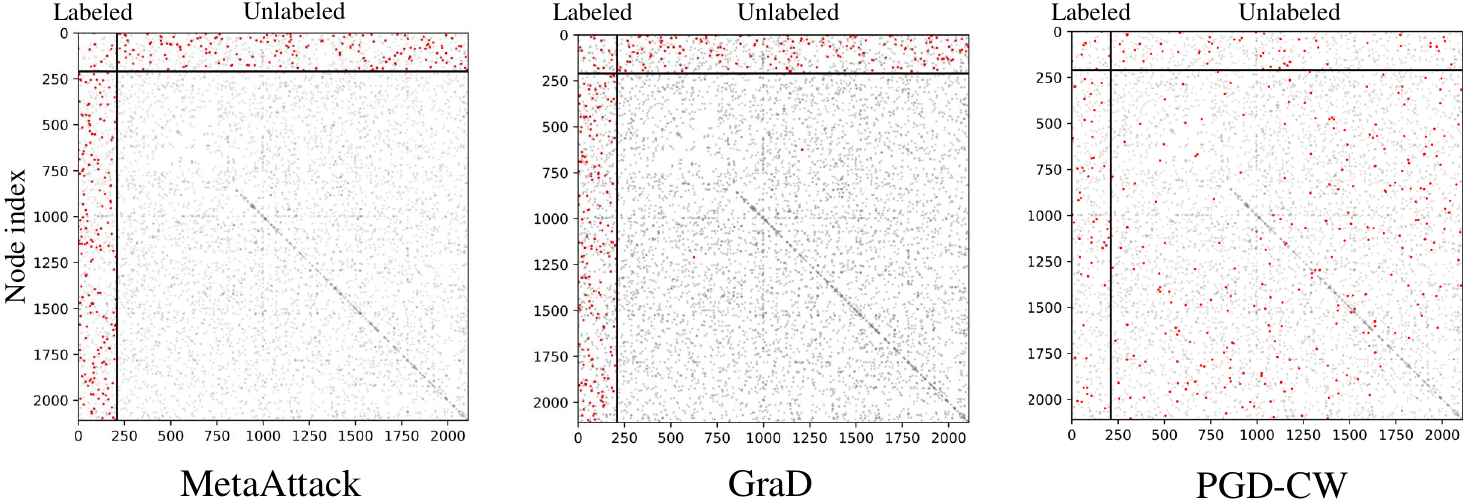}
    \Description[<short description>]{<long description>}
    \caption{The adjacency matrix of 5\% perturbed Citeseer dataset. Red represents the flipped edges. The gray represents the original edges. }
    \label{fig:intro}
    \vspace{-4ex}
\end{figure}

However, we find that the meta-gradient-based attacks exhibit a \emph{bias towards the training nodes}. More precisely, this bias manifests as an uneven perturbation of the graph's edges between the training nodes (i.e., labeled nodes) and the other nodes (i.e., unlabeled nodes).
Table~\ref{tab:intro} and Fig~\ref{fig:intro} show the number/ratio of attacks and the adjacency of attacked graphs when meta-gradient-based attacks (e.g., MetaAttack, EpoAtk, AtkSE, and GraD), and PGD\footnote[1]{PGD attacks flip edges based on projected gradient without meta-gradient, where PGD-CE and PGD-CW denote the PGD attacks using the negative cross entropy, and Carlini-Wagner loss \cite{ieee2017-Carlini017-CW} as the attack loss, respectively.} attacks are applied.  We observe that meta-gradient-based attacks tend to connect two nodes when at least one of them is a labeled node (i.e., L-L or L-U\footnote[2]{Note that \textbf{L-U} is the set of edges that connect a \textbf{L}abeled node and an \textbf{U}nlabeled node. \textbf{L-L} and \textbf{U-U} are similarly defined.}), while it is unlikely to connect two unlabeled nodes (i.e., U-U). 
However, these biased attacks are sub-optimal as they do not consider flipping edges within the U-U set at all. This means that even if there exists an edge in the U-U set that could significantly alter the representation of a node when flipped, conventional meta-gradient-based attacks miss this case due to the bias towards training nodes. Considering that only a small subset of nodes in the graph are labeled (i.e., training nodes), excluding the expansive U-U edge set from potential attack targets is not an effective strategy.
Moreover, we find that such uneven perturbations in the graph structure is specific to meta-gradients-based attack methods, while PGD attacks that flip edges without the use of meta-gradients do not exhibit similar trends. We argue that understanding this intriguing phenomenon is crucial for the development of a general attack framework, yet it remains largely unexplored. 

In this paper, we focus on investigating the meta-gradients to uncover the root cause of the uneven perturbations of existing meta-gradient-based attacks.
Specifically, we find that the root cause stems from the inherent design of the meta-gradient-based attack methods.
This is particularly due to the training procedure of the surrogate model, which is solely trained on the training nodes. We call it  \emph{the bias towards training nodes} of the meta-gradient (Sec.~\ref{section:meta-anal}). Based on our findings, we propose a new attack method called \textbf{\textsf{Meta}} attack using the \textbf{\textsf{con}}trastive surrogate objective (\proposed), which alleviates the bias in meta-gradient using a new surrogate loss that includes both labeled (training) and unlabeled nodes in the training procedure (Sec.~\ref{section:metacon}). Finally, we conduct extensive experiments to verify that~\proposed~is more effective in attacking the graph by searching edges to be flipped from L-L, L-U, and U-U compared with existing meta-gradient-based attacks (Sec.~\ref{section:experiment}).

Our main contributions are summarized as follows:
\begin{itemize}[leftmargin=7.5mm]

    \item We find that existing meta-gradient-based attacks show unevenly distributed perturbations between training nodes (i.e., labeled nodes) and the other nodes (i.e., unlabeled nodes), which stems from the inherent design of the meta-gradient-based attacks regarding the training procedure of the surrogate model.

    \item We propose a new attack method called~\proposed~that alleviates the bias in meta-gradients using two contrastive surrogate objectives: the \textit{sample contrastive} and the \textit{dimension contrastive} losses. We provide the theoretical substantiation that the goal of these objectives closely aligns with that of the target GNNs, which the attack model aims to deteriorate.
    
    \item We conduct extensive experiments to demonstrate that~\proposed~outperforms existing meta-gradient-based methods by considering the entire edges as candidates for attack.
\end{itemize}

\section{Related Works}

\subsection{Adversarial Attacks on GNNs}
\looseness=-1
Adversarial attacks on GNNs aim to disrupt the prediction of GNN models by giving small perturbations to the graph in either the training or test phase. There are attack methods that modify node features or graph structure~\cite{iclr18-mettack}, inject new nodes~\cite{www20-node-injection-nipa}, or rewire edges~\cite{kdd21-rewire-attack}.

\looseness=-1
\subsubsection{Targeted and Untargeted attack} Adversarial attacks on graphs can be categorized based on the attacker’s goal. In targeted attacks, the attacker specifically selects the nodes for which the predictions are to be disturbed. RL-S2V~\cite{pmlr-rl2sv-advattack-on-graph} employs a reinforcement learning approach to generate a targeted attack using the prediction of labels. Nettack~\cite{18kdd-nettack} modifies both node features and the graph structure, while preserving the important characteristics of a graph (e.g., co-occurrence of node features and the degree distribution). 
In the untargeted attack, the attacker tries to deteriorate the performance of GNNs across the entire nodes in a graph. The mainstream untargeted attack methods utilize \emph{the meta-gradient}, which is a gradient of the attack loss with respect to the adjaceny matrix. Specifically, MetaAttack~\cite{iclr18-mettack} first introduces the meta-gradient to decide which edges to flip.
Although the meta-gradient is useful to find edges for attacking the graph, MetaAttack depends on the gradient with the largest absolute value, which however is prone to error due to the discrete nature of graphs. To resolve this issue, EpoAtk~\cite{icdm20-epoatk} adopts an exploratory strategy, and AtkSE~\cite{cikm22-atkse} employs graph augmentations and the momentum update to stabilize the meta-gradient. 
Moreover, GraD \cite{liu2022grad_towards} addresses the bias of attack methods that typically flip edges connected nodes with low-confident predictions, which are already misclassified. Hence, GraD proposes a new attack loss that primarily flips edges connected to correctly classified nodes.
Another line of untargeted attack methods uses the projected gradient descent (PGD) to perturb the graph structure. PGD and min-max attacks~\cite{ijacai19-pgd-topology-attack-defense} solve a relaxed combinatorial optimization problem to perturb the graph structure. 
The meta-gradient-based attacks are widely adopted to execute the poisoning attack, rather than PGD attacks due to the destructive power of meta-gradient~\cite{li2023revisiting}.

\subsubsection{Attacker's Knowledge}
Graph adversarial attacks can be categorized by the attacker's knowledge. The untargeted attack methods have been studied in the gray-box setting, where the attacker's knowledge is restricted to input data (i.e., node features and adjacency matrix) and training labels, while the parameters of the victim model are unknown. 
While most existing attack methods are under the gray-box setting, \cite{mujkanovic2022are-defense} recently demonstrates that the adaptive attack can severely degrade the robust GNN methods under white-box attack, where the attacker can access full information, including the input data, the training labels, and the model parameters. On the other hand, some works present a black-box setting \cite{chang2020restricted,ma2021practical}, which is the most practical, where only input data is known to the attacker, while training labels are unknown.


In this paper, we investigate the uneven perturbation of the graph structure observed in untargeted attack methods that employ meta-gradients, as illustrated in Fig~\ref{fig:intro}. Following previous studies~\cite{iclr18-mettack,cikm22-atkse,liu2022grad_towards}, we demonstrate the attack methods under the gray-box setting. Our work is closely related to \cite{li2023revisiting}, which also explores a similar phenomenon but through the lens of distribution shift. 
While their study primarily focuses on elucidating the destructive power of existing meta-gradient-based attack methods,
the work lacks providing a comprehensive understanding of the root cause of the uneven perturbation. On the other hand, our work identifies the \emph{bias towards training nodes} of meta-gradients as the root cause of this phenomenon. Based on the understanding of the bias, we argue that alleviating the bias can enhance the efficacy of meta-gradient-based attack methods, since existing attack approaches have only confined their attack search space to the area surrounding training nodes, i.e., L-L and L-U. To resolve the issue caused by the bias, we propose a new attack method, called~\proposed, which alleviates the bias towards training nodes by using a contrastive surrogate objective.

\section{Problem Formulation}
\looseness=-1
We follow the common gray-box setting in adversarial attacks on graphs~\cite{iclr18-mettack}. 
Let $\mathcal{G}= \langle \mathcal{V},\mathcal{E} \rangle$ denote  a graph, where $\mathcal{V}=\{v_1,...,v_n\}$ is the set of nodes and $\mathcal{E}\in \mathcal{V}\times \mathcal{V}$ is the set of edges. Given a graph $\mathcal{G}$ that is associated with the adjacency matrix $\mathbf{A}\in \{0, 1\}^{n\times n}$ and node features $\mathbf{X}\in \mathbb{R}^{n\times d}$, where $n$ is the number of nodes, and $d$ is the dimension of node features,  we aim to solve the semi-supervised node classification task in which a GNN classifier $f_{\theta}(\mathbf{A}, \mathbf{X})$ is trained to predict the class label $\mathbf{Y}\in\mathbb{R}^{n\times K}$ of nodes, where $K$ is the number of classes.
In the semi-supervised setting, we have the set of labeled nodes $\mathcal{V}_L$, i.e., training nodes, and the set of unlabeled nodes $\mathcal{V}_U$, where $\mathcal{V}=\mathcal{V}_L\cup\mathcal{V}_U$, $\mathcal{V}_L\cap\mathcal{V}_U=\emptyset$ and $|\mathcal{V}_L| 	\ll |\mathcal{V}_U|$. 
To generate an adversarially attacked graph $\tilde{\mathbf{A}}$ that degrades the performance of the GNN classifier $f_\theta$, we solve the following discrete combinatorial optimization problem:
\begin{equation}
\small
    \min_{\tilde{\mathbf{A}}} \mathcal{L}_{\text{atk}}(f_{\theta^{*}}(\tilde{\mathbf{A}}, \mathbf{X}), \hat{\mathbf{Y}})  \quad\quad \text{s.t.} \quad \norm{\tilde{\mathbf{A}}-\mathbf{A}}_{0} \leq \Delta
    \label{eq:co}
\end{equation}
\looseness=-1
where $\hat{\mathbf{Y}}\in \mathbb{R}^{n\times K}$ is the concatenation of the labels of the training nodes in $\mathcal{V}_L$ and the pseudo-labels of the unlabeled nodes in $\mathcal{V}_U$, which are estimated by the pre-trained GNN classifer $f_{\theta^{*}}$.
$\Delta$ is the perturbation budget for the imperceptibility constraint of structure attacks. A common choice for $\mathcal{L}_{\text{atk}}$ is the negative cross-entropy. By flipping edges that greatly decrease $\mathcal{L}_{\text{atk}}$, the attacker expects that the victim model fails to generalize to the attacked graph during either the training or the testing phase. The nodes used to compute the attack loss can be selected by the attacker among labeled nodes, unlabeled nodes, or both the labeled nodes and unlabeled nodes.

\section{On Meta-gradient-based Attack}
\label{section:meta-anal}
Recall that in Sec.~\ref{section:intro} we described our findings that meta-gradient-based attacks unevenly perturb the graph structure between labeled and unlabeled nodes. 
In this section, we revisit the meta-gradient-based attack and investigate the root cause of the unevenly distributed perturbations through a theoretical analysis and empirical studies.

\begin{figure*}[ht!]
    \centering
    \includegraphics[width=1.78\columnwidth]{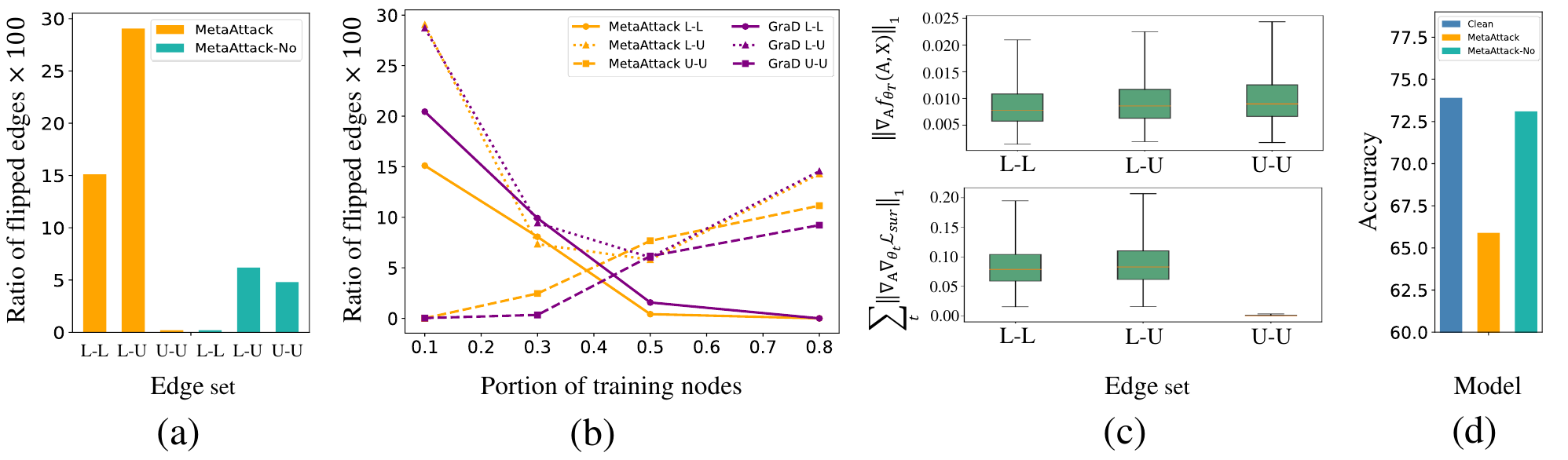}
    \vspace{-1.5ex}
    \Description[<short description>]{<long description>}
    \caption{(a) The ratio of flipped edges for each edge set. A large value means attacks are biased. (b) The ratio of flipped edges over various portions of training nodes. (c) $L_1$ norm visualization of (i) $\nabla_{\mathbf{A}} f_{\theta_T}(\mathbf{A}, \mathbf{X})$ and (ii) $\sum_{t=1}^{T} \nabla_{\mathbf{A}}\nabla_{\theta_{t}} \mathcal{L}_{\text{sur}}$ when $T=10$ in Eqn.~\ref{eq:meta-grad}. (d) Node classification accuracy of MetaAttack and MetaAttack-No. Citeseer dataset is used for the experiments.}
    \label{fig:investigate_meta}
    \vspace{-1.5ex}
\end{figure*}

\subsection{Revisiting Meta-gradient-based Attack}
Since solving Eqn.~\ref{eq:co} is a NP-hard problem, it is difficult to obtain the optimally attacked graph that deteriorates the performance of GNN. To resolve the issue, \cite{iclr18-mettack} introduces MetaAttack that determines which edges to flip based on the meta-gradient of the attack loss with respect to the adjacency matrix (i.e., $\nabla_{\mathbf{A}}\mathcal{L}_{\text{atk}}$). Specifically, it solves the following optimization problem: 
\begin{align}
    \min_{\tilde{\mathbf{A}}} \mathcal{L}_{\text{atk}}(f_{\theta^{*}}(\tilde{\mathbf{A}}, &\mathbf{X}), \hat{\mathbf{Y}}) \nonumber 
    \,\,\, \text{s.t.} \,\,\,\, \theta^{*} = \arg\min_{\theta} \mathcal{L}_{\text{sur}}(f_{\theta}(\tilde{\mathbf{A}}, \mathbf{X}), \mathbf{Y}_L) \label{eq:bilvl_optim} \\
    & \text{where} \quad \norm{\tilde{\mathbf{A}}-\mathbf{A}}_{0} \leq \Delta 
\end{align}
\noindent where at each iteration the outer loop flips a single edge with the largest absolute value of meta-gradient to minimize the attack loss, while the inner loop trains the surrogate model by using the modified graph. MetaAttack alternatively updates the inner and outer loop. 
{The negative cross entropy and the cross entropy losses are used for $\mathcal{L}_{\text{atk}}$ and $\mathcal{L}_{\text{sur}}$, respectively (i.e., $\mathcal{L}_{\text{atk}}=-\mathcal{L}_{\text{sur}}$).}
Note that $\mathcal{L}_{\text{atk}}$ and $\mathcal{L}_{\text{sur}}$ are different in their goals. 

In summary, $\mathcal{L}_{\text{atk}}$ aims to compute the meta-gradient $\nabla_{\mathbf{A}}\mathcal{L}_{\text{atk}}$ that determines edge flips in the outer loop using $\hat{\mathbf{Y}}$, while $\mathcal{L}_{\text{sur}}$ aims to train the surrogate model in the inner loop using only the training nodes $\mathbf{Y}_L$.

\subsection{Investigating Meta-gradient} 
In this section, we analyze the root cause of the uneven perturbations of meta-gradient-based attacks by investigating the meta-gradient.
By unrolling the training procedure, the meta-gradient can be rewritten as follows~\cite{iclr18-mettack}:
{\small
\begin{align}
    & \nabla_{\mathbf{A}} \mathcal{L}_{\text{atk}} = \nabla_f \mathcal{L}_{\text{atk}} \cdot \{\nabla_{\mathbf{A}} f_{\theta_T}  + \nabla_{\theta_T} f_{\theta_T} \cdot \nabla_{\mathbf{A}} \theta_{T} \} \nonumber \\
    &= \nabla_f \mathcal{L}_{\text{atk}} \cdot \{ \underbrace{\nabla_{\mathbf{A}} f_{\theta_T}}_{\text{\textbf{(i)}}} + \nabla_{\theta_T} f_{\theta_T} \cdot (\nabla_{\mathbf{A}}\theta_0 - \underbrace{\alpha \sum_{t=1}^{T} \nabla_{\mathbf{A}}\nabla_{\theta_{t}} \mathcal{L}_{\text{sur}})}_{\text{\textbf{(ii)}}} \}
    \label{eq:meta-grad}
    \raisetag{55pt}
\end{align}
}%
where $T$ is the number of iterations for training the surrogate model in Eqn. \ref{eq:bilvl_optim}, {$\alpha$ is the learning rate}, and $\theta_{t}$ is the parameter of surrogate model after performing gradient descent $t$ times. Herein, we abuse the notation $f_{\theta}(\mathbf{A}, \mathbf{X})$ as $f_{\theta}$. Derivation of Eqn. \ref{eq:meta-grad} can be found in~\cite{online_appendix}.
It is important to note that the meta-gradient involves the training procedure of the surrogate model by treating $\theta$ as a variable without fixing it. We would like to emphasize that two terms in Eqn. \ref{eq:meta-grad} are important for determining the value of the meta-gradient\footnote[3]{{$\nabla_{\theta_{T}} f_{\theta_T}(\mathbf{A}, \mathbf{X})$ and $\nabla_{\mathbf{A}} \theta_0$ do not contribute to determining which edge to flip. $\nabla_{\theta_{T}} f_{\theta_T}(\mathbf{A}, \mathbf{X})$ is the output with the fixed $\mathbf{A}$, which is not related to the change of $\mathbf{A}$, and $\nabla_{\mathbf{A}} \theta_0$ is the gradient of the initial parameter of the surrogate model, which is a constant.}}: \textbf{(i)} how greatly the prediction of the GNN classifier $f_\theta$ changes, and \textbf{(ii)} the gradients of the training loss of the surrogate model. Specifically, the meta-gradient is large when the flipped edge in $\mathbf{A}$ greatly changes the predicted class probability, which is determined by the first term (i.e., \textbf{(i)}). 
Hence, the meta-gradient-based attack connects nodes with different {features or} labels by decreasing the feature smoothness and the assortativity of the graph \cite{kdd20-prognn}. 
In addition, the second term (i.e., \textbf{(ii)}) implies that the meta-gradient is related to the training procedure of the surrogate model performed in the inner loop. 
Herein, it is important to emphasize that $\mathcal{L}_{\text{sur}}$, i.e., cross-entropy, involves only the training nodes, and this is what causes the uneven perturbations described in Sec.~\ref{section:intro}. In other words, we argue that the design of the meta-gradient-based attack that only involves the training nodes in the training procedure of the surrogate model is the root cause of the uneven perturbations of existing attack methods. Hence, we call the second term (i.e., \textbf{(ii)}) ``the \emph{bias towards training nodes}'' of the meta-gradient.

\subsubsection{Empirical Study} 
To verify whether the bias towards training nodes in MetaAttack indeed induces the uneven perturbation, we conduct several studies. Note that we run MetaAttack with the learning rate $\alpha=0.01$ under the common semi-supervised node classification setting with the $10\%$ training portion. We select the negative cross-entropy loss computed on unlabeled nodes as attack loss i.e., $\mathcal{L}_{\text{atk}}$ as it shows the best attack performance. Due to the page limit, we report the result with the above setting. We have observed a similar tendency when using other attack losses.

\noindent {\emph{Empirical Study 1:}} \@ A naive way to verify the impact of the bias towards training nodes on the uneven perturbation phenomenon is to remove the training procedure of the surrogate model in the inner loop, i.e., remove \textbf{(ii)}. More precisely, we pretrain the surrogate model and then fix the parameters to $\theta^*$ thereafter. We name this method MetaAttack-No. In Fig \ref{fig:investigate_meta}.a, we observe that MetaAttack-No has a larger ratio of flipped edges for U-U than MetaAttack, while it has a smaller ratio for L-L and L-U. This clearly shows that removing the bias caused by the training procedure of the surrogate model is beneficial for alleviating the trend of mainly attacking the training nodes, implying that the bias is related to the uneven perturbation phenomenon.

\noindent {\emph{Empirical Study 2:}} \@ Another way to verify the relationship between the bias towards training nodes and the uneven perturbation is to adjust the bias (i.e., \textbf{(ii)}) by changing the portion of training nodes. 
In other words, if we increase the small portion of training nodes to a larger portion, the bias towards the training nodes will be mitigated. 
In Fig~\ref{fig:investigate_meta}.b, we generally observe that the ratio of the flipped edges for U-U increases as we increase the number of training nodes for training the surrogate model, while that for L-L and L-U decreases. The results can be interpreted as follows: With a smaller training portion, $\mathcal{L}_{\text{sur}}$ is computed on a small number of training nodes. This means there are fewer edge perturbations that can alter $\mathcal{L}_{\text{sur}}$. This leads to a tendency to mainly flip edges surrounding the training nodes, i.e., L-L and L-U. On the other hand, as the training portion increases, there are more edge perturbations that can alter $\mathcal{L}_{\text{sur}}$. 
In this situation, the \textbf{(i)} term becomes more influential in determining the largest absolute value of the meta-gradient, while the impact of the \textbf{(ii)} term diminishes. This alleviates the tendency to flip edges between training nodes. These findings suggest that $\mathcal{L}_{\text{sur}}$ is closely related to the uneven perturbation. In particular, the tendency to flip edges around training nodes is caused by the training procedure of the surrogate loss, which is exclusively computed on the training nodes.

Moreover, in Fig~\ref{fig:investigate_meta}.c, we visualize the $L_1$ norm of the two terms in Eqn.~\ref{eq:meta-grad}, i.e., \textbf{(i)} and \textbf{(ii)}, respectively, which are important for determining the value of the meta-gradient. 
We observe that $\norm{\nabla_{\mathbf{A}}f_{\theta_T}(\mathbf{A}, \mathbf{X})}_1$ is similar in all L-L, L-U, and U-U, which implies that even if the meta-gradient flips an edge that greatly changes the prediction probability, it is not associated with the uneven perturbation between labeled and unlabeled nodes. On the other hand, $\sum_{t=1}^{T}\norm{\nabla_{\mathbf{A}}\nabla_{\theta_t} \mathcal{L}_{\text{sur}}}_1$ of L-L and L-U is significantly larger than that of U-U, which implies that the training procedure of the surrogate model is indeed the main factor for the uneven perturbation between labeled nodes and unlabeled nodes. 
As existing meta-gradient-based attack methods, such as EpoAtk~\cite{icdm20-epoatk}, AtkSE~\cite{cikm22-atkse}, and GraD~\cite{liu2022grad_towards}, include the training procedure of the surrogate model in the meta-gradient computation, they inevitably suffer from the bias towards training nodes.
Recall that the uneven perturbation phenomenon was alleviated in MetaAttack-No (See Fig~\ref{fig:investigate_meta}.a), since $\sum_{t=1}^{T}\norm{\nabla_{\mathbf{A}}\nabla_{\theta_t} \mathcal{L}_{\text{sur}}}_1$ term is not included in the meta-gradient computation when the inner loop is removed. 
However, it is important to note that although this naive solution alleviates the bias, the effectiveness of the attack is greatly reduced as shown in Fig~\ref{fig:investigate_meta}.d, i.e., the drop in the node classification accuracy after applying MetaAttack-No is smaller than MetaAttack.

In summary, we found out that the root cause of the uneven perturbation phenomenon in which nodes around the training nodes are mainly attacked is \emph{the training prodecure of the surrogate model}
in the meta-gradient-based attack, i.e., \emph{the bias towards training nodes}. 

We argue that while including the training procedure of the surrogate model in the meta-gradient computation is helpful for attacking the GNN models, the bias towards training nodes confine the search space of potential attacking edges around training nodes. 
That is, even if there exists an edge in the U-U set that could significantly alter the representation of a node when flipped, conventional meta-gradient-based attacks miss this case due to the bias towards training nodes, which causes their suboptimal performance.

\section{Bias-mitigated Meta-gradient-based Attack}
\label{section:metacon}

In this section, we propose \textbf{\textsf{Meta}} attack with \textbf{\textsf{Con}}trastive surrogate objective (\textbf{\textsf{Metacon}}), a novel approach to meta-gradient-based attacks that addresses the bias towards training nodes.
We begin by outlining our research goal and the challenges \textbf{(Sec. \ref{subsec:method_overview})}. Subsequently, we elaborate on two contrastive surrogate objectives of~\proposed, \textit{the sample contrastive loss} and \textit{the dimension contrastive loss}. We provide a theoretical analysis to support the use of these contrastive objectives in surrogate model training \textbf{(Sec. \ref{subsec:method_contraobj})}. 
Finally, we assess the computational complexity of meta-gradient-based attack methods to highlight the efficiency of~\proposed~\textbf{(Sec. \ref{subsec:method_complexity})}. 
Note that~\proposed~can be applied to any attack methods, since it solely requires changing the surrogate objective of attack models. 

\subsection{Overview}
\label{subsec:method_overview}

\noindent \textbf{Goal.} \@ Building on our prior analysis, we have determined that the utilization of meta-gradients is crucial for achieving strong attack performance. However, the inherent bias within meta-gradients results in suboptimal performance in meta-gradient-based attacks. 
Therefore, our primary focus is on alleviating the bias of meta-gradient in order to address the constrained search space issue prevalent in existing meta-gradient-based attack methods. 
To this end, we propose a bias-alleviated attack method, which effectively impairs the performance of victim GNNs. This is achieved by expanding the attack search space to a broader range of edge sets, not just limited to L-L and L-U edges.

\noindent \textbf{Challenges.} \@ To mitigate the bias present in meta-gradient-based attacks, it becomes crucial to incorporate the influence of both labeled and unlabeled nodes in the surrogate model's training procedure. We recognize that introducing a surrogate loss that encompasses all nodes can help alleviate the bias towards training nodes.
One straightforward approach is to use the loss function computed on both labeled and unlabeled nodes, such as the link reconstruction loss~\cite{kipf2016graphvae}. However, simply incorporating the unlabeled nodes in the surrogate loss falls short of generating effective attacks if the goal of the loss (i.e., link reconstruction) does not align with that of the victim GNNs (i.e., the node classification). This will be demonstrated in Table~\ref{tab:effect_surrogate}. 
Therefore, we aim to design an attack model using the surrogate objective, which incorporates unlabeled nodes while still aligning with the task objectives of the victim GNNs.

        

    

\subsection{Meta Attack via Contrastive Surrogate Objective}
\label{subsec:method_contraobj}

Since accessing the ground truth labels of unlabeled nodes is typically not feasible, it is challenging to incorporate these nodes into \textit{supervised} surrogate losses, such as cross-entropy loss. An alternative approach can be employing \textit{self-supervised} surrogate losses, whose goal closely aligns with that of the cross entropy loss, as will be proven in this section. Here, we propose the two following surrogate objectives for~\proposed: the \textit{sample contrastive} and the \textit{dimension contrastive} surrogate losses. 
Subsequently, we offer theoretical substantiation to reinforce the validity of introducing these objectives.

\subsubsection{Sample Contrastive Surrogate Objective (\textsf{Metacon-S})} 

Drawing inspiration from self-supervised graph representation learning~\cite{www21-gca}, we first introduce the sample contrastive surrogate loss for unlabeled nodes. Formally, let $p$ and $\hat{p}$ represent the output probabilities of the surrogate models $f_{\theta}(\mathbf{A}, \mathbf{X})$ and $f_{\theta}(\hat{\mathbf{A}}, \hat{\mathbf{X}})$, respectively, where $\hat{\mathbf{A}}$ and $\hat{\mathbf{X}}$ denote the augmented adjacency matrix and node feature matrix, respectively.
We retain the original graph without any modifications, and generate a single augmented graph view. Based on the two output probabilities $p$ and $\hat{p}$, the sample contrastive surrogate loss on node $u$ is computed as:
{\small
\begin{equation}
    l_\text{s-con}(u) = -\log \frac{e^{s(p_u, \hat{p}_u)}}{e^{s(p_u, \hat{p}_u)} + \sum_{k=1}^n \left[ \mathds{1}_{\left[ k\neq u\right]}e^{s(p_u, \hat{p}_k)} + \mathds{1}_{\left[ k\neq u\right]}e^{s(\hat{p}_u, \hat{p}_k)} \right] }
    \label{eq:samcontrastive_surro}
\end{equation}}
where $s$ is the cosine similarity function. The sample contrastive loss encourages the output probabilities of the same node $u$ in the original graph and augmented graph to be close, while making the output probabilities of different nodes (i.e., samples) in the original graph and augmented graph distant from one another. Consequently, the final surrogate loss of~\proposed~is $\mathcal{L}_{\text{sur}} = \mathcal{L}_\text{ce} + \lambda \mathcal{L}_\text{s-con}$, where $\mathcal{L}_\text{ce}$ is the cross-entropy loss computed on labeled nodes, and $\mathcal{L}_\text{s-con}$ is computed on unlabeled nodes i.e., $\mathcal{L}_\text{s-con}=\sum_{u\in\mathcal{V}_U} {l}_\text{s-con}(u)$. Optimizing the revised surrogate loss alleviates the bias towards training nodes by incorporating the unlabeled nodes into the loss. Note that while the contrastive loss can also be optimized for training nodes~\cite{khosla2020supervised}, our primary objective is to involve unlabeled nodes in the training procedure of the surrogate model to mitigate the bias towards training nodes; thus, we exclude the training nodes from $\mathcal{L}_\text{s-con}$. Moreover, we empirically observed that including the training nodes into $\mathcal{L}_\text{s-con}$ only increases the model complexity without improving the attack performance of~\proposed.

We further justify employing the proposed surrogate loss $\mathcal{L}_{\text{sur}}$, which includes the contrastive loss computed on unlabeled nodes, through theoretical analysis. Specifically, \textit{the goal of surrogate loss computed on unlabeled nodes should align closely with that of surrogate loss} computed on training nodes, i.e., the cross-entropy loss $\mathcal{L}_\text{ce}$. We give the rationality of the surrogate loss of~\proposed~through the following theorem:

\begin{theorem}\label{theorm:surrogate}
Assume that a surrogate model $f_\theta (\cdot)$ consists of multiple GCN layers without non-linear activation, and the mean of the probabilities obtained from $p_u$ and $\hat{p}_u$ is the same. When the node classes are balanced, $\mathcal{L}_{\text{s-con}}$ is the upper bound of the cross entropy. 
\end{theorem}
\looseness=-1

The proof can be found in~\cite{online_appendix}. The theorem indicates that minimizing $\mathcal{L}_{\text{s-con}}$ is approximately minimizing the cross entropy loss. That is, the goal of the contrastive surrogate loss applied on the unlabeled nodes aligns with that of the cross entropy surrogate loss applied on the labeled nodes. Note that the attack methods may not effectively degrade the victim model's performance if the goals of surrogate loss computed on labeled nodes and unlabeled nodes are different. 
For instance, we can consider adding the link prediction loss, which is also unsupervised, instead of the contrastive loss to include more of unlabeled nodes in the training procedure of the surrogate model. 
However, it would not be as effective an attack as the contrastive loss-based surrogate model, because the goal of the link prediction loss does not align with the contrastive loss, and thus cross-entropy loss, which we demonstrate in Sec.~\ref{subsection:experiment-ablation}.

\subsubsection{Dimension Contrastive Surrogate Objective (\textsf{Metacon-D})} 

While using the sample contrastive loss for training the surrogate model offers advantages in generating bias-mitigated attacks, it also presents a notable computational cost, particularly when computing the similarities between all nodes in the denominator of Eqn.~\ref{eq:samcontrastive_surro}. 
Hence, we introduce another attack method, which effectively reduces the computational cost by utilizing the dimension contrastive surrogate loss. Formally, let $\mathbf{P}=f_{\theta}(\mathbf{A}, \mathbf{X})\in \mathbb{R}^{n \times K}$ and $\mathbf{\hat{P}}=f_{\theta}(\mathbf{\hat{A}}, \mathbf{\hat{X}})\in \mathbb{R}^{n \times K}$ be the probability matrices of entire $n$ nodes in the original graph and an augmented graph, respectively. Then, the dimension contrastive loss~\cite{bardes2022vicreg, garrido2023on_dual_con} can be computed as:
\begin{equation}
    \mathcal{L}_{\text{d-con}} =  {m}(\mathbf{P}, \hat{\mathbf{P}}) + \mu_1 (v(\mathbf{P})+v(\hat{\mathbf{P}})) + \mu_2 (c(\mathbf{P})+ c(\hat{\mathbf{P}})) \\
    \label{eq:dim_contrastive_surro}
\end{equation}
where $\mu_1$ and $\mu_2$ are hyperparameters for each criterion. Specifically, $m(\mathbf{P}, \hat{\mathbf{P}}) = \frac{1}{n} \sum_{u \in \mathcal{V}} ||p_u - \hat{p}_u||_2^2$ represents the average of mean-squared euclidean distance, which reduces the distance between probability vectors in the original view and the augmented view. $v(\mathbf{P}) = \frac{1}{K} \sum_{j=1}^K \sqrt{\max \{0, \gamma - Var(p^j)) \}}$ is the hinge loss for the variance of the probability matrix, where $p^j\in\mathbb{R}^n$ is the vector composed of each value at the $j^\text{th}$ dimension in
all row vectors in $\mathbf{P}$,
and $Var(p^j)$ is the scalar variance of the element at the $j^\text{th}$ dimension computed across all nodes. This criterion forces the variance of all node probability vectors to be $\gamma$ along each dimension, preventing the trivial solution where all the node map on the same probability vector. $c(\mathbf{P}) = \frac{1}{K} \sum_{i\neq j} [C(\mathbf{P})]_{i,j}^2$ is the sum of the off-diagonal elements of covariance matrix $C(\mathbf{P})$, where $C(\mathbf{P})=\frac{1}{n-1} \sum_{u \in \mathcal{V}} (p_u-\bar{p})(p_u-\bar{p})^T$. Minimizing this criterion makes the probability vectors sharpened for a specific class. By contrasting the node representation in a dimension-wise manner, dimension contrastive learning is more computationally efficient compared to sample contrastive learning, which contrasts all pairs of nodes.

Now, we need to confirm whether the use of the dimension contrastive loss aligns with the goal of the surrogate model trained with the cross entropy loss. 
We provide the validity based on a recent study that has demonstrated the connection between the sample contrastive learning and the dimension contrastive learning~\cite{garrido2023on_dual_con}. Specifically, \cite{garrido2023on_dual_con} has shown that the objectives of sample contrastive and dimension contrastive learning are closely related. To be more precise, they are equivalent up to row and column normalization of the representation matrix of a batch. This implies that the dimension contrastive loss can be replaced with the sample contrastive loss of~\proposed~without compromising the task objectives of the victim GNN. For the proof of the equivalence between two contrasitve losses, please refer to~\cite{online_appendix}. Similar to how the sample contrastive loss is employed as a surrogate loss, we incorporate the dimension contrastive loss into our attack method by optimizing the final surrogate loss, $\mathcal{L}_{\text{sur}} = \mathcal{L}_\text{ce} + \lambda \mathcal{L}_\text{d-con}$. We refer to the attack methods that utilize the surrogate model trained with $\mathcal{L}_\text{s-con}$ and $\mathcal{L}_\text{d-con}$ as \textsf{Metacon-S} and \textsf{Metacon-D}, respectively.

\subsubsection{Greedy Selection Attack}
Following the design of MetaAttack,~\proposed~alternatively trains the surrogate model during $T$ iterations, and computes the meta-gradient $\nabla_{\mathbf{A}}\mathcal{L}_{\text{atk}}$ based on the surrogate model to decide which edge to flip, where $\mathcal{L}_{\text{atk}}$ is the negative cross-entropy. Note that we only apply the contrastive loss to the surrogate loss $\mathcal{L}_{\text{sur}}$, not to the attack loss $\mathcal{L}_{\text{atk}}$.

\subsection{Discussion on Complexity}
\label{subsec:method_complexity}

Consider a graph with $n$ nodes and $m$ edges. Both MetaAttack and~\proposed~employ a GCN with the same number of parameters, resulting in a computational cost of $\mathcal{O}(n+m)$ during a forward pass~\cite{thakoor2022bgrl}. For MetaAttack, the time complexity for backpropagation is similar to that of the forward pass. This process is repeated for $T$ inner training iterations when perturbing a single edge, leading to a total time complexity of $\mathcal{O}((n+m)T)$. On the other hand, \textsf{Metacon-S} employs the sample contrastive loss, which considers all nodes other than the target node itself as negatives, resulting in $n^2$ loss computations in the denominator term. Consequently, the total time complexity for \textsf{Metacon-S} amounts to $\mathcal{O}((n^2+nm)T)$. Conversely, \textsf{Metacon-D} only requires $n K^2$ loss computation, resulting in a total time complexity of $\mathcal{O}((nK^2+mK^2)T)$. 

Considering that the squared dimension size of the output probability vector $K^2$ is significantly smaller than $n$, i.e., $K^2  \ll  n$, \textsf{Metacon-D} demonstrates computational efficiency compared to \textsf{Metacon-S}. Moreover, it incurs only $K^2$ times the computational complexity of MetaAttack, making it a computationally affordable option. Please refer to a more detailed analysis of the complexity in Sec.~\ref{subsec:complexity}.

\section{Experiment}
\label{section:experiment}

We design extensive experiments to answer the following research questions. \textbf{(RQ1)} Does~\proposed~effectively alter the graph structure compared to existing graph attack methods? \textbf{(RQ2)} Does~\proposed~resolve the confined attack search space problem of meta-gradient-based methods? Lastly, we analyze components in~\proposed~, its hyperparameters, and complexity.

\subsection{Experimental Setup}

\subsubsection{Datasets}
We evaluate~\proposed~and baselines on the benchmark datasets, such as citation networks, i.e., Cora, Cora ML, Citeseer, a social network, i.e., Polblogs and Reddit, and co-purchase network, i.e., amazon photo and amazon computers. Following the experimental setup of~\cite{iclr18-mettack}, we use the largest connected component of the graph. We randomly split into 10\%/10\%/80\% for training, validation, and testing, respectively. The statistics of the datasets are shown in Table~\ref{tab:dataset}.

\begin{table}[t!]
\centering
\caption{Statistics of datasets.}
\label{tab:dataset}

{\small
\begin{tabular}{c|c|cccc}

 Dataset & \# Nodes & \# Edges & \# Features & \# Classes \\
\hline 
\hline
 Cora     & 2,485          & 5,069         & 1,433      & 7   \\  
 Cora ML     & 2,810          & 7,981         & 2,879      & 7   \\  
 Citeseer   & 2,110        & 3,668         & 3,703      & 6   \\
 Polblogs   & 1,222        & 16,714       & 0        &  2   \\
 Am. Photo   & 7,650        & 119,081       & 745        &  8   \\
 Am. Computers   & 13,752        & 245,861       & 767        &  10   \\
 Reddit   & 231,443        & 11,606,919       & 602        &  41   \\
\hline

\end{tabular}

}
\vspace{0ex}
\end{table}

\begin{table}[t]
\centering
\caption{Node classification accuracy under the strong transfer scenario. 5\% of edges are flipped on Cora, Citeseer, Polblogs and Cora ML datasets. \textbf{Bold} represents the best performance. \underline{Underline} represents the second place.}
\label{table:main1}
\scalebox{.80}{
\begin{tabular}{ll|cccc}
\toprule
\multicolumn{2}{c|}{Dataset}  & Cora & Citeseer  & Polblogs & Cora ML \\ \midrule

\multicolumn{2}{c|}{Clean} & 83.6±0.3 & 73.9±0.5 & 95.0±0.8 & 85.3±1.1 \\ \midrule

\multirow{2}{*}{\textsf{Rand}}
& \text{Random} & 82.7±0.2	& 73.3±0.8 & 91.6±1.2	& 84.0±1.2 \\
& \text{DICE} & 82.3±0.6	&  73.1±1.0   &   89.7±0.3	  &   84.2±0.9  \\ \midrule

\multirow{2}{*}{\textsf{Self}}   
& BBGA & 82.7±0.5 & 73.4±1.2 & 87.7±0.4 & 84.9±0.7   \\
& CLGA  & 81.2±0.3 & 72.4±1.1 & 88.2±1.4 & 84.8±0.7   \\ \midrule

\multirow{2}{*}{\textsf{PGD}} & 

PGD-CE & 83.7±0.6 & 73.3±0.7 & 83.5±0.6 & 85.1±0.6 \\
& PGD-CW & 80.6±0.7 & 70.9±0.8 & 78.2±1.6 & 81.7±0.9\\ \midrule

\multirow{6}{*}{\textsf{Meta}} 
& MetaAttack & 76.9±0.6 & 65.9±1.3 & 76.6±0.5 & 76.4±1.3 \\
& EpoAtk  & 82.9±0.3 & 73.0±1.4 & 94.4±0.5 & 84.6±1.0  \\
& AtkSE  & 79.5±2.3 & 72.0±0.9 & 78.7±1.1 & 80.6±1.4    \\ 
& GraD  & 76.8±2.4 & 66.4±2.0 & \textbf{75.1±0.9} & {76.1±1.1}  \\  \cmidrule{2-6}
& \textsf{Metacon-S}  & \underline{75.4±1.5} & \underline{64.1±0.7} & \textbf{75.1±0.5} & \underline{76.0±1.2}  \\  
& \textsf{Metacon-D} & \textbf{75.3±1.1} & \textbf{63.9±0.8} & {75.2±0.6} & \textbf{75.7±0.8}  \\  \bottomrule
\end{tabular}
}

\end{table}

\subsubsection{Baselines} 
We use several untargeted attack methods as baselines, and divide them into four categories: The basic attack using randomness (\textsf{Rand}), the transferred attack from self-supervised learning models without label information (\textsf{Self}), PGD attack (\textsf{PGD}), and meta-gradient-based attacks (\textsf{Meta}). Note that the direct baselines are \textsf{Meta}, but we include \textsf{Self} and \textsf{PGD} to highlight that they cannot be the solution for mitigating the bias towards training nodes. 

\begin{itemize}[leftmargin=4mm]
    \item \textsf{Rand}: \textbf{Random} randomly adds or deletes edges. \textbf{DICE}~\cite{WaniekMRW16-dice} randomly adds edges between nodes with different labels, and deletes edges between nodes with the same labels.

    \item \textsf{Self}: This approach attacks GNNs without label information. \textbf{CLGA} \cite{www22-clga-attack} uses a self-supervised graph representation learning model \cite{grace,www21-gca}) as a surrogate model to generate an attacked graph. \textbf{BBGA} \cite{ml2022-dealing-uneven} performs spectral clustering to obtain pseudo-labels for all nodes, and then attacks the GCN surrogate model trained with the pseudo-labels. We transfer the obtained graph from both CLGA and BBGA to a new victim model.

    \item \textsf{PGD}: This approach utilizes the projected gradient descent~\cite{ijacai19-pgd-topology-attack-defense} to attack the graph. We employ two attack losses, the negative cross entropy and the CW loss. 
    
    \item \textsf{Meta}: This approach contains meta-gradient-based methods, which includes \textbf{MetaAttack}\footnote[5]{Among several variants of MetaAttack, we choose Meta-Self in \cite{iclr18-mettack} as a baseline, where the attack loss is computed on unlabeled nodes using pseudo-label.}, and its extensions. \textbf{Epoatk} \cite{icdm20-epoatk} performs MetaAttack with an exploration strategy. \textbf{Atkse}~\cite{cikm22-atkse} stabilizes the meta-gradients with the graph augmentation. \textbf{GraD}~\cite{liu2022grad_towards} encourages the attack methods to flip edges near correctly classified nodes with the debiased attack loss. 
\end{itemize}

\begin{table*}[t]
    \begin{minipage}{1.17\columnwidth}
        \caption{Node classification accuracy under the weak transfer scenario. 5\% of edges are flipped on Cora, Citeseer, Polblogs and Cora ML datasets. \textbf{Bold} represents the best performance. \underline{Underline} represents the second place.}
        \label{table:main2}
        \vspace{-2.5ex}
        \scalebox{0.6}{
        \begin{tabular}{ll|cc|cc|cc|cc}
        \toprule
        \multicolumn{2}{c|}{Dataset}  & \multicolumn{2}{c|}{Cora} & \multicolumn{2}{c|}{Citeseer}  & \multicolumn{2}{c|}{Polblogs} & \multicolumn{2}{c}{Cora ML} \\ 
        
        \multicolumn{2}{c|}{Victim Model}  & GraphSAGE & GAT & GraphSAGE & GAT & GraphSAGE & GAT & GraphSAGE & GAT \\ \midrule
        
        \multicolumn{2}{c|}{Clean} & 81.6±1.6& 83.8±0.6 & 72.7±1.1 & 73.5±0.8 & 95.1±1.0 & 94.9±0.3 & 84.5±1.0 & 85.2±0.8 \\ \midrule
        
         \multirow{2}{*}{\textsf{Self}} & BBGA  & 82.3±0.6 & 82.2±0.8 & 72.4±0.9 & 73.7±0.6 & 93.3±0.7 & 92.1±0.9 & 84.4±0.4 & 84.3±0.7 \\
         & CLGA   & 80.7±0.8 & 81.3±0.5 & 71.8±0.4 & 72.6±0.6 & 92.6±1.1& 90.0±1.3 & 82.6±0.6 & 82.5±0.9  \\ \midrule
        \multirow{2}{*}{\textsf{PGD}} & 
        PGD-CE   & 82.7±0.8 & 83.8±0.5 & 73.3±1.0 & 74.1±0.8 & \textbf{85.6±0.4} & 83.5±1.0 & 85.1±0.4 & 85.5±0.8  \\
        & PGD-CW  & 80.5±0.7 & 80.7±0.9 & 71.4±0.9 & 70.9±1.0 & 87.1±0.8 & \textbf{79.8±1.2} & 82.6±0.5 & 81.5±1.0 \\ \midrule 
        \multirow{6}{*}{\textsf{Meta}} & MetaAttack & 78.2±1.1 & 78.7±0.9  & 69.4±1.0 & 69.1±0.0  & \underline{86.6±1.8} & \underline{80.9±0.9} & 80.9±1.2 & 79.8±1.2\\
        & EpoAtk  & 81.6±1.2 & 82.7±0.4 & 72.4±0.5 & 73.4±0.6 & 94.5±0.6& 94.5±0.3 & 84.2±0.3 & 84.3±0.8 \\
        & AtkSE  &  80.5±1.1 & 81.9±1.2 & 72.4±1.3 & 73.9±0.8 & 91.7±1.4 & 88.8±3.0 & 82.4±1.2 & 81.5±1.4 \\ 
        & GraD  & 78.5±0.8 & 79.1±1.0 & 69.3±1.8 & 69.2±1.0 & 86.9±0.5 & {81.2±0.7} & {80.7±1.1} & \textbf{78.6±1.4} \\  \cmidrule{2-10}
        & \textsf{Metacon-S}  & \textbf{77.3±1.0} & \textbf{77.0±1.1} & \textbf{66.4±2.1} & \underline{68.0±0.9} & 86.8±1.3 & 81.5±1.4 & \underline{80.4±1.0} & \underline{78.8±1.4} \\  
        & \textsf{Metacon-D}  & \underline{77.9±0.7} & \textbf{77.0±1.0} & \underline{66.6±1.9} & \textbf{66.8±1.2} & {86.9±1.3} & 82.1±0.8 & \textbf{80.2±0.5} & {78.9±0.9} \\  
        \bottomrule
        \end{tabular}
        }
        
    \end{minipage}
    \hfill
    \begin{minipage}{0.9\columnwidth}
        \caption{Node classification accuracy against robust GNN models. $5\%$ edges are perturbed.}
        \label{table:defense}
        \vspace{-2ex}
        \scalebox{0.64}{
        \begin{tabular}{l|ccc|ccc}
        \toprule
        \multicolumn{1}{c|}{Dataset}  & \multicolumn{3}{c|}{Cora} & \multicolumn{3}{c}{Citeseer} \\ \midrule
        
        \multicolumn{1}{c|}{Victim Model}  & GCN-JCD & RGCN & ProGNN & GCN-JCD & RGCN & ProGNN \\ \midrule\midrule
        
        \multicolumn{1}{c|}{Clean} & 79.1±0.5 & 81.6±1.6 & 83.8±0.6 &  71.8±1.1 & 73.5±0.8 & 73.3±0.7 \\ \midrule\midrule
        
        MetaAttack & 77.4±0.7 & 75.6±0.6	& 83.0±0.4	&  70.9±0.4	& 66.4±0.5	& 72.5±0.7	\\
        GraD  & 77.6±0.6     &  75.9±0.3	& 83.3±0.6		& 	70.8±0.5 & 66.5±0.5	& 72.9±0.3	 \\  \midrule
        \textsf{Metacon-S}  &  77.8±0.7    & {75.5±0.4} & \textbf{82.8±0.2} &  70.8±0.6 & \textbf{65.1±0.4} & \textbf{71.7±0.4}  \\  		
        \textsf{Metacon-D}  &  \textbf{77.2±0.7}    & \textbf{74.2±0.5} & 
        {83.1±0.4} & \textbf{70.7±0.8}  & {66.3±0.8} & {73.1±0.5}  \\  		
        \bottomrule
        \end{tabular}
        }
        
        \vspace{1ex}
        \caption{Effect of the surrogate loss.}
        \label{tab:effect_surrogate}
        \vspace{-2ex}
        \scalebox{0.62}{
        \begin{tabular}{c|cccc}
        \hline
         Model & Cora & Citeseer & Polblogs & Cora ML \\
        \hline 
        \hline
         MetaAttack ($\mathcal{L}_\text{sur} = \mathcal{L}_\text{ce}$)& 76.9±0.6 & 65.9±1.3 & 76.6±0.5 & 76.4±1.3 \\
         MetaAttack w/ $\mathcal{L}_\text{link}$ ($\mathcal{L}_\text{sur} = \mathcal{L}_\text{ce} + \lambda\mathcal{L}_{\text{link}}$)   &  77.9±0.7 & 70.2±0.4 & 75.8±0.7 & 76.9±2.0 \\
         \textsf{Metacon-S}($\mathcal{L}_\text{sur} = \mathcal{L}_\text{ce} + \lambda\mathcal{L}_{\text{s-con}}$) & {75.4±1.5} & {64.1±0.7} & \textbf{75.1±0.5} & {76.0±1.2}  \\  
         \textsf{Metacon-D}($\mathcal{L}_\text{sur} = \mathcal{L}_\text{ce} + \lambda\mathcal{L}_{\text{d-con}}$) & \textbf{75.3±1.1} & \textbf{63.9±0.8} & {75.2±0.6} & \textbf{75.7±0.1}  \\  
        \hline
        \end{tabular}
        }
        
    \end{minipage}

\end{table*}

\subsubsection{Evaluation Protocol} We perform the node classification task given an attacked graph, where $\Delta= \#\text{ }\textsf{edges}\times \textsf{perturbation ratio}$, and the perturbation ratios are $5\%, 10\%, 15\%$, and $20\%$. We consider the poisoning attacks (i.e., training time attack) to evaluate \proposed~and baselines. Specifically, a \textit{poisoning attack} perturbs the adjacency matrix before the model is trained. i.e., a victim model is trained with an attacked graph.

Following existing studies~\cite{iclr18-mettack, cikm22-atkse}, we choose 2-layer GCN models without ReLu activation for the surrogate model to generate attacked graphs. Then, we consider two scenarios: 1) The strong transfer, which disrupts the GCN victim model with the poisoned graph. 2) The weak transfer, which disrupts the GraphSAGE and GAT victim models.

\subsubsection{Implementation Details} 

We implement baselines using their official codes with the hyperparameters given in the paper. For hyperparameters of~\proposed, we tune $\lambda$ and the edge dropping ratio for augmented graph by grid search with an interval of 0.1 from 0.0 to 1.0. For hyperparameters of dimension contrastive loss, we fix $\gamma=1$ without searching it, and only tune $\mu_1$ and $\mu_2$ in the range of $\{1.0, 2.0, 4.0\}$. We report the average and standard deviation of 5 runs. We used a A6000 GPU device with 48GB memory for each experiment.

\subsection{Comparison with Graph Attack Methods}
\label{subsection:experiment-comparison}

\subsubsection{Strong Transfer Attack}

\begin{figure}[t!]
    \centering
    \Description[]{}
    \includegraphics[width=0.9\columnwidth]{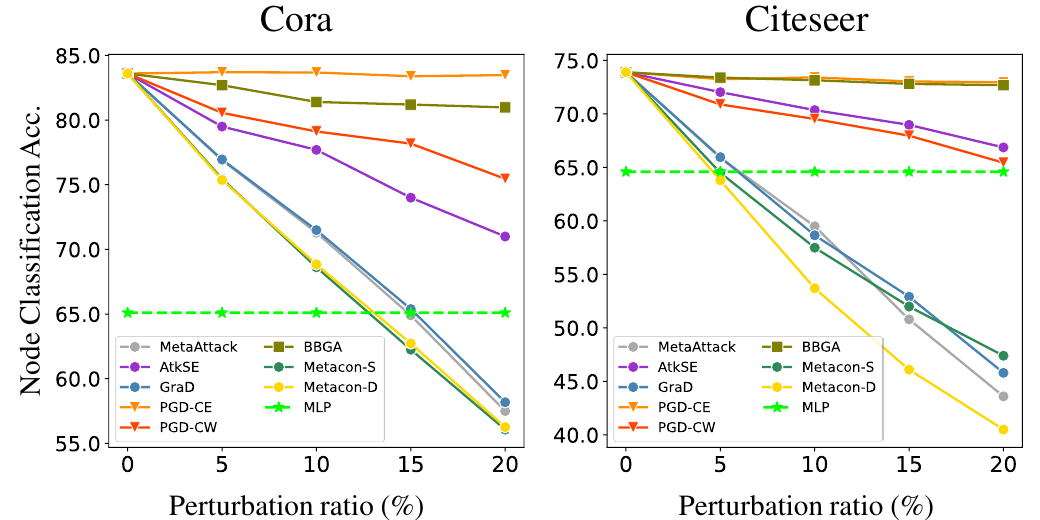}
    \caption{Node classification accuracy over perturbation ratios under the strong transfer scenario. }
    \label{fig:acc_by_ptbrate}
    \vspace{-3ex}
\end{figure}

Table \ref{table:main1} shows the result of the node classification task under an attacked graph, where $5\%$ of edges in Cora, Citeseer, Polblogs, and Cora ML datasets are perturbed. We have the following observations: \textbf{1)} Attack methods in \textsf{Meta} generally outperform the baseline attack methods in \textsf{Self}, and \textsf{PGD}, which do not utilize the meta-gradient. This shows that meta-gradients are effective in attacking graph. Moreover, although attack methods in \textsf{Self} and \textsf{PGD} produce bias-mitigated attacks, they cannot be the solution for mitigating bias due to their ineffective attack performance.
\textbf{2)} Among the meta-gradient-based attacks, \textsf{Metacon-S} and \textsf{Metacon-D} greatly degrade the performance of the GCN victim model, implying that alleviating the bias towards the training node is helpful for generating effective attacks to the GCN model. Similarly, in Fig~\ref{fig:acc_by_ptbrate}, we observe that our proposed methods achieves the state-of-art attack performance over various perturbation ratios. 

\subsubsection{Weak Transfer Attack}

Table \ref{table:main2} shows the node classification task under an attacked graph when victim models are GraphSAGE and GAT, respectively. We observe that \textsf{Metacon-S} and \textsf{Metacon-D} achieve the state-of-art attack performance on Cora and Citeseer in terms of degrading the node classification accuracy, while showing comparable performances on Cora ML. This implies that the bias-alleviated attacked graph generally degrades the performance of various GNN models besides GCN.

\begin{figure}[t!]
    \centering
    \Description[]{}
    \includegraphics[width=0.9\columnwidth]{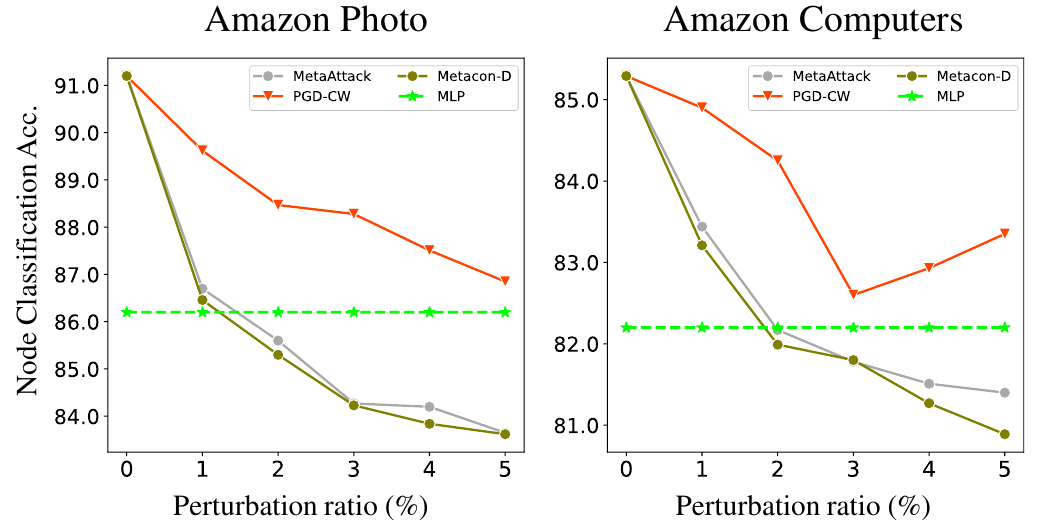}
    \caption{Node classification accuracy over perturbation ratios on large-scale networks under the strong transfer scenario. Amazon Photo and Computers datasets are used.}
    \label{fig:acc_by_ptbrate_largescale}
\end{figure}

\subsubsection{Transfer Attack to Robust GNN Models}

In Table~\ref{table:defense}, we evaluated the attack methods against various robust GNN models, i.e., GCN-Jaccard~\cite{ijcai19-advexamples-for-graph}, RGCN~\cite{kdd19-rgcn}, and ProGNN~\cite{kdd20-prognn}. We observed that \textsf{Metacon-S} and \textsf{Metacon-D} effectively attack the graph structure for not only conventional GNNs but also state-of-the-art defense models. It implies that the attack model, which considers U-U edges as a candidate, is more difficult to defend for the robust GNN models compared to biased attacks such as MetaAttack and GraD.

\subsubsection{Evaluation on Large-scale Graphs}
As meta-gradient-based methods (i.e., MetaAttack, GraD, and~\proposed) require substantial memory for training, we conducted experiments on larger graphs by sampling subgraphs~\cite{www2022ariel}. Specifically, we divided the graphs into subgraphs, each containing a maximum of 3,000 nodes, and then generated attacks on each subgraph. We finally added all attacked edges to the original entire graph. 
In Fig~\ref{fig:acc_by_ptbrate_largescale} and Table~\ref{table:reddit}, we observed that~\proposed~achieves competitive performance with the meta-gradient-based methods, showing the effectiveness of considering U-U edges when attacking graphs on large-scale networks.

\begin{table}[!t]
\centering
\vspace{-2.2ex}
\caption{Node Classification Accuracy against GCN models. $0.05\%$ of edges are perturbed on Reddit.}
\scalebox{0.78}{
\begin{tabular}{l|cccc}
\toprule
\multicolumn{1}{c|}{Attack Model}  & Clean & MetaAttack & \textsf{Metacon-S} &\textsf{Metacon-D} \\ \midrule

\multicolumn{1}{c|}{Acc} & 95.17±0.04 & 95.04±0.04 & OOM & \textbf{94.94±0.06}  \\ 
\bottomrule
\end{tabular}
}
\vspace{-2ex}
\label{table:reddit}
\end{table}

\subsection{Analysis on Bias Alleviation}
\label{section:experiment-bias}
\begin{figure}[t]
    \centering
    \Description[]{}
    \includegraphics[width=0.75\columnwidth]{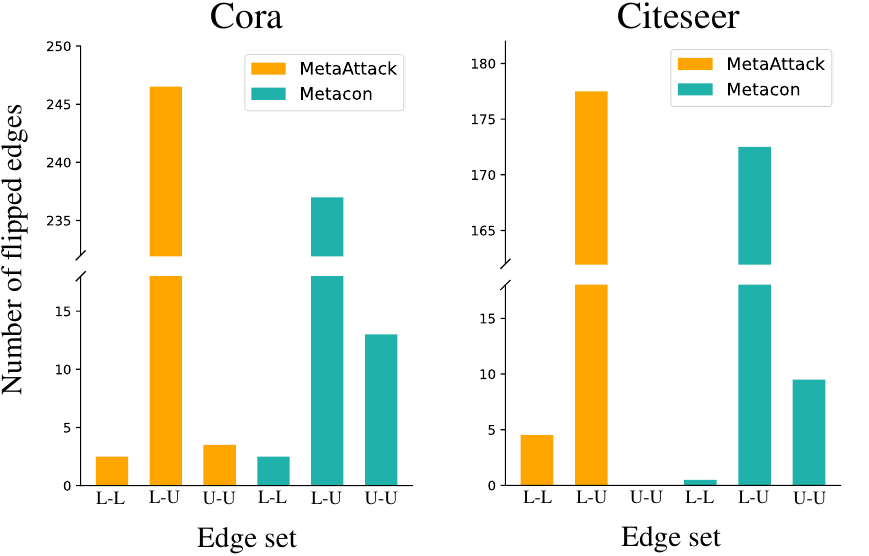}
    \vspace{-1ex}
    \caption{Number of flipped edges on Cora and Citeseer datasets when MetaAttack and~\textsf{Metacon-S} are applied.}
    \label{fig:bias_alleviation}
    \vspace{-3ex}
\end{figure}

We compare~\proposed~with MetaAttack to analyze the attacking trend in terms of uneven perturbations. Fig~\ref{fig:bias_alleviation} shows the number of attacks for L-L, L-U, and U-U, respectively. We observe that~\proposed~increases the number of attacks for U-U, while reducing that for L-L and L-U compared to MetaAttack. This implies that adding the proposed contrastive surrogate loss computed on the unlabeled nodes in addition to the cross-entropy loss computed on the training nodes expands the search space of meta-gradient-based attack methods, thereby improving the attack performance. This result demonstrates that the proposed surrogate loss alleviates the bias towards training nodes.

\subsection{Ablation Study}
\label{subsection:experiment-ablation}

\subsubsection{Effect of Surrogate loss}

We further conduct ablation studies to investigate the effectiveness of the proposed surrogate loss under the strong transfer scenario. First, we remove the contrastive surrogate loss by setting $\lambda=0$, i.e., $\mathcal{L}_{\text{sur}}= \mathcal{L}_\text{ce}$, which is equivalent to MetaAttack. Second, we replace the contrastive surrogate loss with the link reconstruction loss, i.e., $\mathcal{L}_{\text{sur}}= \mathcal{L}_\text{ce} + \lambda \mathcal{L}_\text{link}$, which reconstructs the edges in the graph. The link reconstruction loss also involves unlabeled nodes in the surrogate loss. In Table~\ref{tab:effect_surrogate}, we have the following observations: \textbf{1)} When the contrastive surrogate loss is removed, i.e., $\mathcal{L}_{\text{sur}}= \mathcal{L}_\text{ce}$, the performance of~\proposed~severely degrades, which demonstrates the effectiveness of the contrastive surrogate loss. \textbf{2)} When MetaAttack uses the surrogate loss with the link reconstruction loss, the node classification accuracy of MetaAttack is increased, implying that the link prediction loss is not helpful for attacking the graph. 
This implies that aligning the goal of the surrogate loss applied on labeled and unlabeled nodes is crucial.

\begin{figure}[t]
    \centering
    \Description[]{}
    \includegraphics[width=0.75\columnwidth]{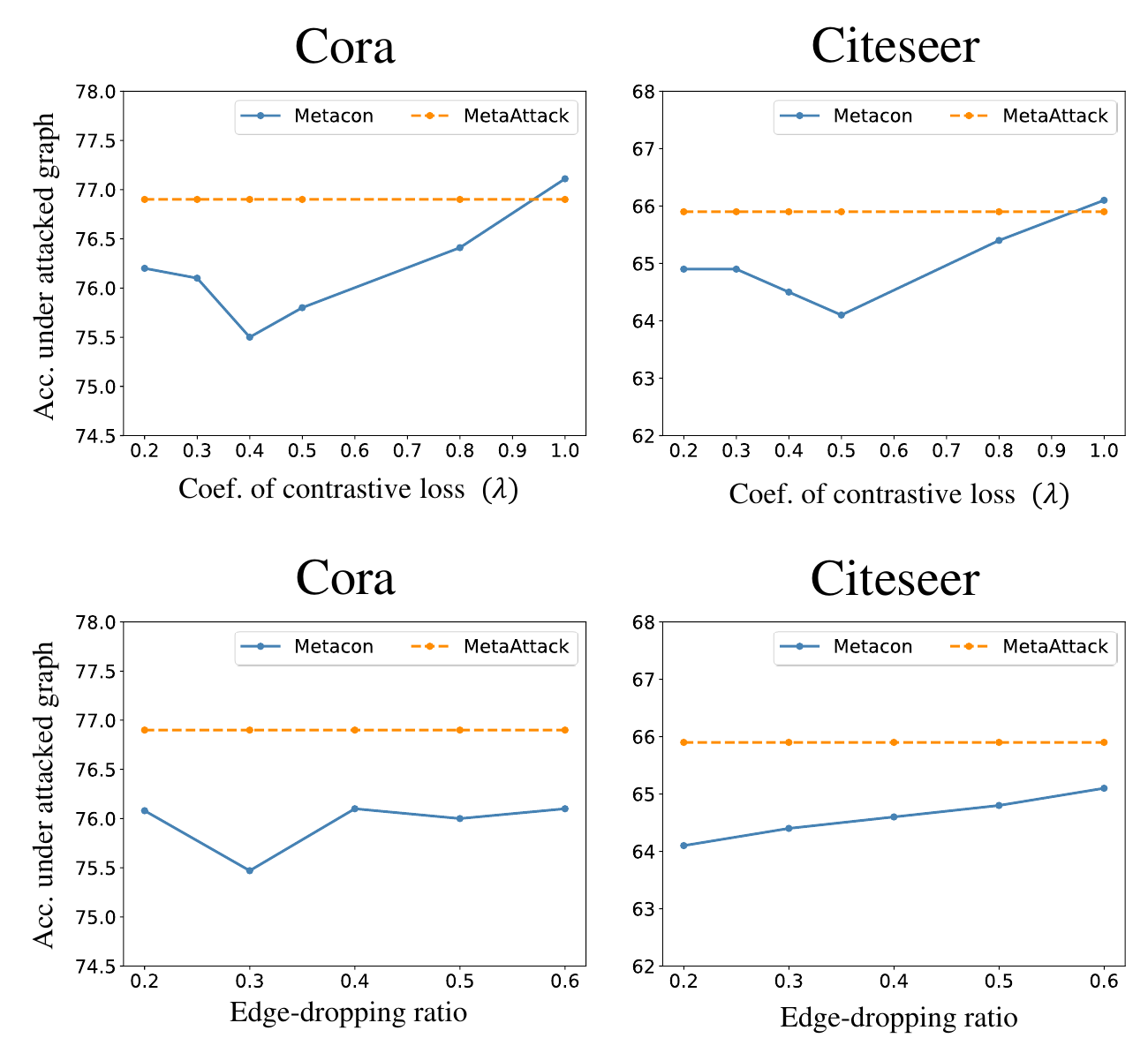}
    \caption{Analysis on $\lambda$ and edge-dropping ratio. The node classification accuracy under the attacked graph, where $5\%$ edges are flipped on Cora and Citeseer datasets.}
    \label{fig:hyper}
    \vspace{-2ex}
\end{figure}

\subsubsection{Hyperparameter} We further analyze the hyperparameter sensitivity. Fig~\ref{fig:hyper} shows the node classification accuracy under an attacked graph when we vary the coefficient of the contrastive loss $\lambda$ and edge-dropping ratio of the augmented graph, respectively. For $\lambda$, we observed that~\proposed~greatly outperforms the attack performance of MetaAttack over various $\lambda$, implying that the contrastive loss of~\proposed~is helpful for improving the attack performance of MetaAttack. For the edge-dropping ratio, we observed that~\proposed~consistently outperforms MetaAttack. This implies that the performance of~\proposed~is not sensitive to the stochastic view generation of the contrastive learning.

\begin{table}[t]

\captionof{table}{Analysis on the complexity of the attack methods. The memories required for the attack models are reported.}
\scalebox{0.8}{
        
        \begin{tabular}{l|cc}
        \toprule
        \multicolumn{1}{c|}{Dataset}  & \multicolumn{1}{c|}{Cora} & \multicolumn{1}{c}{Citeseer} \\ \midrule
        
         \# Nodes & 2,485 & 2,110 \\ 
         \# Edges & 5,069 & 3,668 \\
         \# Feats & 1,433 & 3,703 \\ \midrule
         MetaAttack & 5.5 GB & 5.8 GB \\
         \textsf{Metacon-S} & 22.4 GB & 23.6 GB \\ 
         \textsf{Metacon-D} & 6.3 GB & 6.5 GB \\ \bottomrule

        \end{tabular}}
        \vspace{-3ex}
        
        \label{table:complexity}
\end{table}

\subsection{Time Complexity}
\label{subsec:complexity}

As we discussed in Sec.~\ref{subsec:method_complexity}, MetaAttack requires the total time complexity of $\mathcal{O}((n+m)T)$. On the other hand, the total time complexity for \textsf{Metacon-S} and \textsf{Metacon-D} amounts to $\mathcal{O}((n^2+nm)T)$ and $\mathcal{O}((nK^2+mK^2)T)$, respectively. Table~\ref{table:complexity} shows the memory usage per single edge perturbation when MetaAttack, \textsf{Metacon-S}, and \textsf{Metacon-D} are used, respectively. The memory usage is proportional to the requirement of gradient computation of each method. We observed that \textsf{Metacon-S} requires 4.1 times the computation of MetaAttack per single edge perturbation, which is very expensive. However, \textsf{Metacon-D} requires only 1.1 times the computation of MetaAttack. This implies that \textsf{Metacon-D} achieves the significant improvement in terms of the attack performance with affordable time overhead.

\section{Conclusion}
In this paper, we discovered the uneven perturbation phenomenon of meta-gradient-based attacks, which mainly flips edges around training nodes, i.e., L-L and L-U, and uncover the root cause in the meta-gradient, i.e., bias towards training nodes. Based on the understanding of phenomenon, we propose~\proposed, which alleviates the bias using contrastive surrogate objective. We demonstrated that~\proposed~effectively degrades the performance of the GNN victim models by considering the entire edges as candidates for attack. 
We believe that our study makes a contribution towards understanding the mechanisms employed by existing successful meta-gradient-based attack methods. However, our study is currently limited to addressing uneven perturbations in the aspect of the attacker. Based on the understanding of the phenomenon, we expect that the defense mechanism can also be improved in the future. 

\section*{Acknowledgement}

This work was supported by the National Research Foundation of Korea(NRF) grant funded by the Korea government(MSIT) (RS-2024-00335098), Ministry of Science and ICT (NRF-2022M3J6A1063021), and by the Institute of Information \& communications Technology Planning \& Evaluation (IITP) grant funded by the Korea government(MSIT) (No.2022-0-00157).

\bibliographystyle{ACM-Reference-Format}
\bibliography{acmart}


\begin{thebibliography}{41}


\ifx \showCODEN    \undefined \def \showCODEN     #1{\unskip}     \fi
\ifx \showDOI      \undefined \def \showDOI       #1{#1}\fi
\ifx \showISBNx    \undefined \def \showISBNx     #1{\unskip}     \fi
\ifx \showISBNxiii \undefined \def \showISBNxiii  #1{\unskip}     \fi
\ifx \showISSN     \undefined \def \showISSN      #1{\unskip}     \fi
\ifx \showLCCN     \undefined \def \showLCCN      #1{\unskip}     \fi
\ifx \shownote     \undefined \def \shownote      #1{#1}          \fi
\ifx \showarticletitle \undefined \def \showarticletitle #1{#1}   \fi
\ifx \showURL      \undefined \def \showURL       {\relax}        \fi
\providecommand\bibfield[2]{#2}
\providecommand\bibinfo[2]{#2}
\providecommand\natexlab[1]{#1}
\providecommand\showeprint[2][]{arXiv:#2}

\bibitem[Bardes et~al\mbox{.}(2022)]%
        {bardes2022vicreg}
\bibfield{author}{\bibinfo{person}{Adrien Bardes}, \bibinfo{person}{Jean Ponce}, {and} \bibinfo{person}{Yann LeCun}.} \bibinfo{year}{2022}\natexlab{}.
\newblock \showarticletitle{{VICR}eg: Variance-Invariance-Covariance Regularization for Self-Supervised Learning}. In \bibinfo{booktitle}{\emph{International Conference on Learning Representations}}. \bibinfo{publisher}{{}}, \bibinfo{address}{{}}, \bibinfo{pages}{{}}.
\newblock
\urldef\tempurl%
\url{https://openreview.net/forum?id=xm6YD62D1Ub}
\showURL{%
\tempurl}


\bibitem[Carlini and Wagner(2017)]%
        {ieee2017-Carlini017-CW}
\bibfield{author}{\bibinfo{person}{Nicholas Carlini} {and} \bibinfo{person}{David~A. Wagner}.} \bibinfo{year}{2017}\natexlab{}.
\newblock \showarticletitle{Towards Evaluating the Robustness of Neural Networks}. In \bibinfo{booktitle}{\emph{2017 {IEEE} Symposium on Security and Privacy, {SP} 2017, San Jose, CA, USA, May 22-26, 2017}}. \bibinfo{publisher}{{IEEE} Computer Society}, \bibinfo{address}{{}}, \bibinfo{pages}{39--57}.
\newblock
\urldef\tempurl%
\url{https://doi.org/10.1109/SP.2017.49}
\showDOI{\tempurl}


\bibitem[Chang et~al\mbox{.}(2020)]%
        {chang2020restricted}
\bibfield{author}{\bibinfo{person}{Heng Chang}, \bibinfo{person}{Yu Rong}, \bibinfo{person}{Tingyang Xu}, \bibinfo{person}{Wenbing Huang}, \bibinfo{person}{Honglei Zhang}, \bibinfo{person}{Peng Cui}, \bibinfo{person}{Wenwu Zhu}, {and} \bibinfo{person}{Junzhou Huang}.} \bibinfo{year}{2020}\natexlab{}.
\newblock \bibinfo{title}{A Restricted Black-box Adversarial Framework Towards Attacking Graph Embedding Models}.
\newblock
\newblock


\bibitem[Dai et~al\mbox{.}(2022)]%
        {wsdm22-rsgnn}
\bibfield{author}{\bibinfo{person}{Enyan Dai}, \bibinfo{person}{Wei Jin}, \bibinfo{person}{Hui Liu}, {and} \bibinfo{person}{Suhang Wang}.} \bibinfo{year}{2022}\natexlab{}.
\newblock \showarticletitle{Towards Robust Graph Neural Networks for Noisy Graphs with Sparse Labels}.
\newblock \bibinfo{journal}{\emph{arXiv preprint arXiv:2201.00232}}  \bibinfo{volume}{{}} (\bibinfo{year}{2022}), \bibinfo{pages}{{}}.
\newblock


\bibitem[Dai et~al\mbox{.}(2018)]%
        {pmlr-rl2sv-advattack-on-graph}
\bibfield{author}{\bibinfo{person}{Hanjun Dai}, \bibinfo{person}{Hui Li}, \bibinfo{person}{Tian Tian}, \bibinfo{person}{Xin Huang}, \bibinfo{person}{Lin Wang}, \bibinfo{person}{Jun Zhu}, {and} \bibinfo{person}{Le Song}.} \bibinfo{year}{2018}\natexlab{}.
\newblock \showarticletitle{Adversarial Attack on Graph Structured Data}. In \bibinfo{booktitle}{\emph{Proceedings of the 35th International Conference on Machine Learning}} \emph{(\bibinfo{series}{Proceedings of Machine Learning Research})}. \bibinfo{publisher}{PMLR}, \bibinfo{address}{{}}, \bibinfo{pages}{{}}.
\newblock


\bibitem[Feng et~al\mbox{.}(2022)]%
        {www2022ariel}
\bibfield{author}{\bibinfo{person}{Shengyu Feng}, \bibinfo{person}{Baoyu Jing}, \bibinfo{person}{Yada Zhu}, {and} \bibinfo{person}{Hanghang Tong}.} \bibinfo{year}{2022}\natexlab{}.
\newblock \bibinfo{title}{ARIEL: Adversarial Graph Contrastive Learning}.
\newblock
\newblock
\showeprint{2208.06956}~[cs.LG]


\bibitem[Garrido et~al\mbox{.}(2023)]%
        {garrido2023on_dual_con}
\bibfield{author}{\bibinfo{person}{Quentin Garrido}, \bibinfo{person}{Yubei Chen}, \bibinfo{person}{Adrien Bardes}, \bibinfo{person}{Laurent Najman}, {and} \bibinfo{person}{Yann LeCun}.} \bibinfo{year}{2023}\natexlab{}.
\newblock \showarticletitle{On the duality between contrastive and non-contrastive self-supervised learning}. In \bibinfo{booktitle}{\emph{The Eleventh International Conference on Learning Representations}}. \bibinfo{publisher}{{}}, \bibinfo{address}{{}}, \bibinfo{pages}{{}}.
\newblock
\urldef\tempurl%
\url{https://openreview.net/forum?id=kDEL91Dufpa}
\showURL{%
\tempurl}


\bibitem[Geisler et~al\mbox{.}(2020)]%
        {neurips20_ReliableAGG}
\bibfield{author}{\bibinfo{person}{Simon Geisler}, \bibinfo{person}{Daniel Z\"{u}gner}, {and} \bibinfo{person}{Stephan G\"{u}nnemann}.} \bibinfo{year}{2020}\natexlab{}.
\newblock \showarticletitle{Reliable Graph Neural Networks via Robust Aggregation}. In \bibinfo{booktitle}{\emph{Advances in Neural Information Processing Systems}}. \bibinfo{publisher}{{}}, \bibinfo{address}{{}}, \bibinfo{pages}{{}}.
\newblock


\bibitem[Haoxi~Zhan(2022)]%
        {ml2022-dealing-uneven}
\bibfield{author}{\bibinfo{person}{Xiaobing~Pei Haoxi~Zhan}.} \bibinfo{year}{2022}\natexlab{}.
\newblock \showarticletitle{Dealing with the unevenness: deeper insights in graph-based attack and defense}.
\newblock \bibinfo{journal}{\emph{Machine Learning}}  \bibinfo{volume}{{}} (\bibinfo{year}{2022}), \bibinfo{pages}{{}}.
\newblock
\showISSN{1573-0565}
\urldef\tempurl%
\url{https://doi.org/10.1007/s10994-022-06234-4}
\showDOI{\tempurl}


\bibitem[In et~al\mbox{.}(2024)]%
        {in2024sggsr}
\bibfield{author}{\bibinfo{person}{Yeonjun In}, \bibinfo{person}{Kanghoon Yoon}, \bibinfo{person}{Kibum Kim}, \bibinfo{person}{Kijung Shin}, {and} \bibinfo{person}{Chanyoung Park}.} \bibinfo{year}{2024}\natexlab{}.
\newblock \showarticletitle{Self-Guided Robust Graph Structure Refinement}. In \bibinfo{booktitle}{\emph{Proceedings of the ACM on Web Conference 2024}} (Singapore, Singapore) \emph{(\bibinfo{series}{WWW '24})}. \bibinfo{publisher}{Association for Computing Machinery}, \bibinfo{address}{New York, NY, USA}, \bibinfo{pages}{697–708}.
\newblock
\showISBNx{9798400701719}
\urldef\tempurl%
\url{https://doi.org/10.1145/3589334.3645522}
\showDOI{\tempurl}


\bibitem[In et~al\mbox{.}(2023)]%
        {in2023spagcl}
\bibfield{author}{\bibinfo{person}{Yeonjun In}, \bibinfo{person}{Kanghoon Yoon}, {and} \bibinfo{person}{Chanyoung Park}.} \bibinfo{year}{2023}\natexlab{}.
\newblock \showarticletitle{Similarity Preserving Adversarial Graph Contrastive Learning}. In \bibinfo{booktitle}{\emph{{}}} (Long Beach, CA, USA) \emph{(\bibinfo{series}{KDD '23})}. \bibinfo{publisher}{Association for Computing Machinery}, \bibinfo{address}{New York, NY, USA}, \bibinfo{pages}{867–878}.
\newblock
\showISBNx{9798400701030}
\urldef\tempurl%
\url{https://doi.org/10.1145/3580305.3599503}
\showDOI{\tempurl}


\bibitem[Jin et~al\mbox{.}(2021a)]%
        {wsdm22-simpgcn}
\bibfield{author}{\bibinfo{person}{Wei Jin}, \bibinfo{person}{Tyler Derr}, \bibinfo{person}{Yiqi Wang}, \bibinfo{person}{Yao Ma}, \bibinfo{person}{Zitao Liu}, {and} \bibinfo{person}{Jiliang Tang}.} \bibinfo{year}{2021}\natexlab{a}.
\newblock \showarticletitle{Node similarity preserving graph convolutional networks}. In \bibinfo{booktitle}{\emph{Proceedings of the 14th ACM International Conference on Web Search and Data Mining}}. \bibinfo{publisher}{{}}, \bibinfo{address}{{}}, \bibinfo{pages}{148--156}.
\newblock


\bibitem[Jin et~al\mbox{.}(2021b)]%
        {kdd21-adv-attack-survey}
\bibfield{author}{\bibinfo{person}{Wei Jin}, \bibinfo{person}{Yaxing Li}, \bibinfo{person}{Han Xu}, \bibinfo{person}{Yiqi Wang}, \bibinfo{person}{Shuiwang Ji}, \bibinfo{person}{Charu Aggarwal}, {and} \bibinfo{person}{Jiliang Tang}.} \bibinfo{year}{2021}\natexlab{b}.
\newblock \showarticletitle{Adversarial Attacks and Defenses on Graphs}.
\newblock \bibinfo{journal}{\emph{{}}} \bibinfo{volume}{22}, \bibinfo{number}{2} (\bibinfo{year}{2021}), \bibinfo{pages}{{}}.
\newblock
\showISSN{1931-0145}
\urldef\tempurl%
\url{https://doi.org/10.1145/3447556.3447566}
\showDOI{\tempurl}


\bibitem[Jin et~al\mbox{.}(2020)]%
        {kdd20-prognn}
\bibfield{author}{\bibinfo{person}{Wei Jin}, \bibinfo{person}{Yao Ma}, \bibinfo{person}{Xiaorui Liu}, \bibinfo{person}{Xianfeng Tang}, \bibinfo{person}{Suhang Wang}, {and} \bibinfo{person}{Jiliang Tang}.} \bibinfo{year}{2020}\natexlab{}.
\newblock \showarticletitle{Graph structure learning for robust graph neural networks}. In \bibinfo{booktitle}{\emph{Proceedings of the 26th ACM SIGKDD International Conference on Knowledge Discovery \& Data Mining}}. \bibinfo{publisher}{{}}, \bibinfo{address}{{}}, \bibinfo{pages}{66--74}.
\newblock


\bibitem[Khosla et~al\mbox{.}(2020)]%
        {khosla2020supervised}
\bibfield{author}{\bibinfo{person}{Prannay Khosla}, \bibinfo{person}{Piotr Teterwak}, \bibinfo{person}{Chen Wang}, \bibinfo{person}{Aaron Sarna}, \bibinfo{person}{Yonglong Tian}, \bibinfo{person}{Phillip Isola}, \bibinfo{person}{Aaron Maschinot}, \bibinfo{person}{Ce Liu}, {and} \bibinfo{person}{Dilip Krishnan}.} \bibinfo{year}{2020}\natexlab{}.
\newblock \showarticletitle{Supervised contrastive learning}.
\newblock \bibinfo{journal}{\emph{Advances in neural information processing systems}}  \bibinfo{volume}{33} (\bibinfo{year}{2020}), \bibinfo{pages}{18661--18673}.
\newblock


\bibitem[Kipf and Welling(2016)]%
        {kipf2016graphvae}
\bibfield{author}{\bibinfo{person}{Thomas~N. Kipf} {and} \bibinfo{person}{Max Welling}.} \bibinfo{year}{2016}\natexlab{}.
\newblock \bibinfo{title}{Variational Graph Auto-Encoders}.
\newblock
\newblock
\showeprint[arxiv]{1611.07308}~[stat.ML]


\bibitem[Kipf and Welling(2017)]%
        {iclr2017kipf-gcn}
\bibfield{author}{\bibinfo{person}{Thomas~N. Kipf} {and} \bibinfo{person}{Max Welling}.} \bibinfo{year}{2017}\natexlab{}.
\newblock \showarticletitle{Semi-Supervised Classification with Graph Convolutional Networks}. In \bibinfo{booktitle}{\emph{International Conference on Learning Representations}}. \bibinfo{publisher}{{}}, \bibinfo{address}{{}}, \bibinfo{pages}{{}}.
\newblock
\urldef\tempurl%
\url{https://openreview.net/forum?id=SJU4ayYgl}
\showURL{%
\tempurl}


\bibitem[Lee et~al\mbox{.}(2023)]%
        {lee2023shiftmol}
\bibfield{author}{\bibinfo{person}{Namkyeong Lee}, \bibinfo{person}{Kanghoon Yoon}, \bibinfo{person}{Gyoung~S. Na}, \bibinfo{person}{Sein Kim}, {and} \bibinfo{person}{Chanyoung Park}.} \bibinfo{year}{2023}\natexlab{}.
\newblock \showarticletitle{Shift-Robust Molecular Relational Learning with Causal Substructure}. In \bibinfo{booktitle}{\emph{Proceedings of the 29th ACM SIGKDD Conference on Knowledge Discovery and Data Mining}} (Long Beach, CA, USA) \emph{(\bibinfo{series}{KDD '23})}. \bibinfo{publisher}{Association for Computing Machinery}, \bibinfo{address}{New York, NY, USA}, \bibinfo{pages}{1200–1212}.
\newblock
\showISBNx{9798400701030}
\urldef\tempurl%
\url{https://doi.org/10.1145/3580305.3599437}
\showDOI{\tempurl}


\bibitem[Li et~al\mbox{.}(2023)]%
        {li2023revisiting}
\bibfield{author}{\bibinfo{person}{Kuan Li}, \bibinfo{person}{Yang Liu}, \bibinfo{person}{Xiang Ao}, {and} \bibinfo{person}{Qing He}.} \bibinfo{year}{2023}\natexlab{}.
\newblock \showarticletitle{Revisiting Graph Adversarial Attack and Defense From a Data Distribution Perspective}. In \bibinfo{booktitle}{\emph{The Eleventh International Conference on Learning Representations}}. \bibinfo{publisher}{{}}, \bibinfo{address}{{}}, \bibinfo{pages}{{}}.
\newblock
\urldef\tempurl%
\url{https://openreview.net/forum?id=dSYoPjM5J_W}
\showURL{%
\tempurl}


\bibitem[Lin et~al\mbox{.}(2022)]%
        {kdd22-spectral-attack}
\bibfield{author}{\bibinfo{person}{Lu Lin}, \bibinfo{person}{Ethan Blaser}, {and} \bibinfo{person}{Hongning Wang}.} \bibinfo{year}{2022}\natexlab{}.
\newblock \showarticletitle{Graph Structural Attack by Perturbing Spectral Distance}. In \bibinfo{booktitle}{\emph{Proceedings of the 28th ACM SIGKDD Conference on Knowledge Discovery and Data Mining}} (Washington DC, USA) \emph{(\bibinfo{series}{KDD '22})}. \bibinfo{publisher}{Association for Computing Machinery}, \bibinfo{address}{New York, NY, USA}, \bibinfo{pages}{989–998}.
\newblock
\showISBNx{9781450393850}
\urldef\tempurl%
\url{https://doi.org/10.1145/3534678.3539435}
\showDOI{\tempurl}


\bibitem[Lin et~al\mbox{.}(2020)]%
        {icdm20-epoatk}
\bibfield{author}{\bibinfo{person}{Xixun Lin}, \bibinfo{person}{Chuan Zhou}, \bibinfo{person}{Hong Yang}, \bibinfo{person}{Jia Wu}, \bibinfo{person}{Haibo Wang}, \bibinfo{person}{Yanan Cao}, {and} \bibinfo{person}{Bin Wang}.} \bibinfo{year}{2020}\natexlab{}.
\newblock \showarticletitle{Exploratory Adversarial Attacks on Graph Neural Networks}. In \bibinfo{booktitle}{\emph{2020 IEEE International Conference on Data Mining (ICDM)}}. \bibinfo{publisher}{{}}, \bibinfo{address}{{}}, \bibinfo{pages}{1136--1141}.
\newblock
\urldef\tempurl%
\url{https://doi.org/10.1109/ICDM50108.2020.00138}
\showDOI{\tempurl}


\bibitem[Liu et~al\mbox{.}(2022a)]%
        {cikm22-atkse}
\bibfield{author}{\bibinfo{person}{Zihan Liu}, \bibinfo{person}{Yun Luo}, \bibinfo{person}{Lirong Wu}, \bibinfo{person}{Siyuan Li}, \bibinfo{person}{Zicheng Liu}, {and} \bibinfo{person}{Stan~Z. Li}.} \bibinfo{year}{2022}\natexlab{a}.
\newblock \showarticletitle{Are Gradients on Graph Structure Reliable in Gray-Box Attacks?}. In \bibinfo{booktitle}{\emph{{}}} (Atlanta, GA, USA) \emph{(\bibinfo{series}{CIKM '22})}. \bibinfo{publisher}{Association for Computing Machinery}, \bibinfo{address}{New York, NY, USA}, \bibinfo{pages}{{}}.
\newblock
\showISBNx{9781450392365}
\urldef\tempurl%
\url{https://doi.org/10.1145/3511808.3557238}
\showDOI{\tempurl}


\bibitem[Liu et~al\mbox{.}(2022b)]%
        {liu2022grad_towards}
\bibfield{author}{\bibinfo{person}{Zihan Liu}, \bibinfo{person}{Yun Luo}, \bibinfo{person}{Lirong Wu}, \bibinfo{person}{Zicheng Liu}, {and} \bibinfo{person}{Stan~Z. Li}.} \bibinfo{year}{2022}\natexlab{b}.
\newblock \showarticletitle{Towards Reasonable Budget Allocation in Untargeted Graph Structure Attacks via Gradient Debias}. In \bibinfo{booktitle}{\emph{Advances in Neural Information Processing Systems}}, \bibfield{editor}{\bibinfo{person}{Alice~H. Oh}, \bibinfo{person}{Alekh Agarwal}, \bibinfo{person}{Danielle Belgrave}, {and} \bibinfo{person}{Kyunghyun Cho}} (Eds.). \bibinfo{publisher}{{}}, \bibinfo{address}{{}}, \bibinfo{pages}{{}}.
\newblock
\urldef\tempurl%
\url{https://openreview.net/forum?id=vkGk2HI8oOP}
\showURL{%
\tempurl}


\bibitem[Ma et~al\mbox{.}(2020)]%
        {ma2021practical}
\bibfield{author}{\bibinfo{person}{Jiaqi Ma}, \bibinfo{person}{Shuangrui Ding}, {and} \bibinfo{person}{Qiaozhu Mei}.} \bibinfo{year}{2020}\natexlab{}.
\newblock \bibinfo{title}{Towards More Practical Adversarial Attacks on Graph Neural Networks}.
\newblock
\newblock


\bibitem[Ma et~al\mbox{.}(2021)]%
        {kdd21-rewire-attack}
\bibfield{author}{\bibinfo{person}{Yao Ma}, \bibinfo{person}{Suhang Wang}, \bibinfo{person}{Tyler Derr}, \bibinfo{person}{Lingfei Wu}, {and} \bibinfo{person}{Jiliang Tang}.} \bibinfo{year}{2021}\natexlab{}.
\newblock \showarticletitle{Graph Adversarial Attack via Rewiring}. In \bibinfo{booktitle}{\emph{Proceedings of the 27th ACM SIGKDD Conference on Knowledge Discovery \& amp; Data Mining}} (Virtual Event, Singapore) \emph{(\bibinfo{series}{KDD '21})}. \bibinfo{publisher}{Association for Computing Machinery}, \bibinfo{address}{New York, NY, USA}, \bibinfo{pages}{1161–1169}.
\newblock
\urldef\tempurl%
\url{https://doi.org/10.1145/3447548.3467416}
\showDOI{\tempurl}


\bibitem[Mujkanovic et~al\mbox{.}(2022)]%
        {mujkanovic2022are-defense}
\bibfield{author}{\bibinfo{person}{Felix Mujkanovic}, \bibinfo{person}{Simon Geisler}, \bibinfo{person}{Stephan G{\"u}nnemann}, {and} \bibinfo{person}{Aleksandar Bojchevski}.} \bibinfo{year}{2022}\natexlab{}.
\newblock \showarticletitle{Are Defenses for Graph Neural Networks Robust?}. In \bibinfo{booktitle}{\emph{Advances in Neural Information Processing Systems}}, \bibfield{editor}{\bibinfo{person}{Alice~H. Oh}, \bibinfo{person}{Alekh Agarwal}, \bibinfo{person}{Danielle Belgrave}, {and} \bibinfo{person}{Kyunghyun Cho}} (Eds.). \bibinfo{publisher}{{}}, \bibinfo{address}{{}}, \bibinfo{pages}{{}}.
\newblock
\urldef\tempurl%
\url{https://openreview.net/forum?id=yCJVkELVT9d}
\showURL{%
\tempurl}


\bibitem[Sun et~al\mbox{.}(2020)]%
        {www20-node-injection-nipa}
\bibfield{author}{\bibinfo{person}{Yiwei Sun}, \bibinfo{person}{Suhang Wang}, \bibinfo{person}{Xianfeng Tang}, \bibinfo{person}{Tsung-Yu Hsieh}, {and} \bibinfo{person}{Vasant Honavar}.} \bibinfo{year}{2020}\natexlab{}.
\newblock \showarticletitle{Adversarial Attacks on Graph Neural Networks via Node Injections: A Hierarchical Reinforcement Learning Approach}. In \bibinfo{booktitle}{\emph{Proceedings of The Web Conference 2020}} (Taipei, Taiwan) \emph{(\bibinfo{series}{WWW '20})}. \bibinfo{publisher}{Association for Computing Machinery}, \bibinfo{address}{New York, NY, USA}, \bibinfo{pages}{673–683}.
\newblock
\urldef\tempurl%
\url{https://doi.org/10.1145/3366423.3380149}
\showDOI{\tempurl}


\bibitem[Thakoor et~al\mbox{.}(2022)]%
        {thakoor2022bgrl}
\bibfield{author}{\bibinfo{person}{Shantanu Thakoor}, \bibinfo{person}{Corentin Tallec}, \bibinfo{person}{Mohammad~Gheshlaghi Azar}, \bibinfo{person}{Mehdi Azabou}, \bibinfo{person}{Eva~L Dyer}, \bibinfo{person}{Remi Munos}, \bibinfo{person}{Petar Veli{\v{c}}kovi{\'c}}, {and} \bibinfo{person}{Michal Valko}.} \bibinfo{year}{2022}\natexlab{}.
\newblock \showarticletitle{Large-Scale Representation Learning on Graphs via Bootstrapping}. In \bibinfo{booktitle}{\emph{International Conference on Learning Representations}}. \bibinfo{publisher}{{}}, \bibinfo{address}{{}}, \bibinfo{pages}{{}}.
\newblock
\urldef\tempurl%
\url{https://openreview.net/forum?id=0UXT6PpRpW}
\showURL{%
\tempurl}


\bibitem[Wang et~al\mbox{.}(2019)]%
        {Wang_2019gcf}
\bibfield{author}{\bibinfo{person}{Xiang Wang}, \bibinfo{person}{Xiangnan He}, \bibinfo{person}{Meng Wang}, \bibinfo{person}{Fuli Feng}, {and} \bibinfo{person}{Tat-Seng Chua}.} \bibinfo{year}{2019}\natexlab{}.
\newblock \showarticletitle{Neural Graph Collaborative Filtering}. In \bibinfo{booktitle}{\emph{Proceedings of the 42nd International ACM SIGIR Conference on Research and Development in Information Retrieval}} \emph{(\bibinfo{series}{SIGIR ’19})}. \bibinfo{publisher}{ACM}, \bibinfo{address}{{}}, \bibinfo{pages}{{}}.
\newblock
\urldef\tempurl%
\url{https://doi.org/10.1145/3331184.3331267}
\showDOI{\tempurl}


\bibitem[Waniek et~al\mbox{.}(2018)]%
        {WaniekMRW16-dice}
\bibfield{author}{\bibinfo{person}{Marcin Waniek}, \bibinfo{person}{Tomasz~P. Michalak}, \bibinfo{person}{Talal Rahwan}, {and} \bibinfo{person}{Michael~J. Wooldridge}.} \bibinfo{year}{2018}\natexlab{}.
\newblock \showarticletitle{Hiding Individuals and Communities in a Social Network}.
\newblock \bibinfo{journal}{\emph{Nature Human Behaviour}}  \bibinfo{volume}{abs/1608.00375} (\bibinfo{year}{2018}), \bibinfo{pages}{{}}.
\newblock
\urldef\tempurl%
\url{http://arxiv.org/abs/1608.00375}
\showURL{%
\tempurl}


\bibitem[Wu et~al\mbox{.}(2019)]%
        {ijcai19-advexamples-for-graph}
\bibfield{author}{\bibinfo{person}{Huijun Wu}, \bibinfo{person}{Chen Wang}, \bibinfo{person}{Yuriy Tyshetskiy}, \bibinfo{person}{Andrew Docherty}, \bibinfo{person}{Kai Lu}, {and} \bibinfo{person}{Liming Zhu}.} \bibinfo{year}{2019}\natexlab{}.
\newblock \showarticletitle{Adversarial Examples for Graph Data: Deep Insights into Attack and Defense}. In \bibinfo{booktitle}{\emph{Proceedings of the Twenty-Eighth International Joint Conference on Artificial Intelligence, {IJCAI-19}}}. \bibinfo{publisher}{International Joint Conferences on Artificial Intelligence Organization}, \bibinfo{address}{{}}, \bibinfo{pages}{{}}.
\newblock


\bibitem[Xu et~al\mbox{.}(2019)]%
        {ijacai19-pgd-topology-attack-defense}
\bibfield{author}{\bibinfo{person}{Kaidi Xu}, \bibinfo{person}{Hongge Che}, \bibinfo{person}{Sijia Liu}, \bibinfo{person}{Pin-Yu Chen}, \bibinfo{person}{Tsui-Wei Weng}, \bibinfo{person}{Mingyi Hong}, {and} \bibinfo{person}{Xue Lin}.} \bibinfo{year}{2019}\natexlab{}.
\newblock \showarticletitle{Topology Attack and Defense for Graph Neural Networks: An Optimization Perspective}. In \bibinfo{booktitle}{\emph{Proceedings of the Twenty-Eighth International Joint Conference on Artificial Intelligence (IJCAI-19)}}. \bibinfo{publisher}{{}}, \bibinfo{address}{{}}, \bibinfo{pages}{{}}.
\newblock


\bibitem[Yoon et~al\mbox{.}(2023)]%
        {hawkes}
\bibfield{author}{\bibinfo{person}{Kanghoon Yoon}, \bibinfo{person}{Youngjun Im}, \bibinfo{person}{Jingyu Choi}, \bibinfo{person}{Taehwan Jeong}, {and} \bibinfo{person}{Jinkyoo Park}.} \bibinfo{year}{2023}\natexlab{}.
\newblock \showarticletitle{Learning Multivariate Hawkes Process via Graph Recurrent Neural Network}. In \bibinfo{booktitle}{\emph{Proceedings of the 29th ACM SIGKDD Conference on Knowledge Discovery and Data Mining}} (Long Beach, CA, USA) \emph{(\bibinfo{series}{KDD '23})}. \bibinfo{publisher}{Association for Computing Machinery}, \bibinfo{address}{New York, NY, USA}, \bibinfo{pages}{5451–5462}.
\newblock
\showISBNx{9798400701030}
\urldef\tempurl%
\url{https://doi.org/10.1145/3580305.3599857}
\showDOI{\tempurl}


\bibitem[Yoon et~al\mbox{.}(2024)]%
        {online_appendix}
\bibfield{author}{\bibinfo{person}{Kanghoon Yoon}, \bibinfo{person}{Yeonjun In}, \bibinfo{person}{Namkyeong Lee}, \bibinfo{person}{Kibum Kim}, {and} \bibinfo{person}{Chanyoung Park}.} \bibinfo{year}{2024}\natexlab{}.
\newblock \showarticletitle{Supplementary Material of Debiased Graph Poisoning attack via Contrastive Surrogate Loss}.
\newblock \bibinfo{journal}{\emph{{}}}  \bibinfo{volume}{{}} (\bibinfo{year}{2024}), \bibinfo{pages}{{}}.
\newblock
\urldef\tempurl%
\url{https://github.com/KanghoonYoon/torch-metacon}
\showURL{%
\tempurl}


\bibitem[Zhang et~al\mbox{.}(2022)]%
        {www22-clga-attack}
\bibfield{author}{\bibinfo{person}{Sixiao Zhang}, \bibinfo{person}{Hongxu Chen}, \bibinfo{person}{Xiangguo Sun}, \bibinfo{person}{Yicong Li}, {and} \bibinfo{person}{Guandong Xu}.} \bibinfo{year}{2022}\natexlab{}.
\newblock \showarticletitle{Unsupervised Graph Poisoning Attack via Contrastive Loss Back-Propagation}. In \bibinfo{booktitle}{\emph{Proceedings of the ACM Web Conference 2022}} (Virtual Event, Lyon, France) \emph{(\bibinfo{series}{WWW '22})}. \bibinfo{publisher}{Association for Computing Machinery}, \bibinfo{address}{New York, NY, USA}, \bibinfo{pages}{1322–1330}.
\newblock
\showISBNx{9781450390965}
\urldef\tempurl%
\url{https://doi.org/10.1145/3485447.3512179}
\showDOI{\tempurl}


\bibitem[Zhang and Zitnik(2020)]%
        {neurips20_gnnguard}
\bibfield{author}{\bibinfo{person}{Xiang Zhang} {and} \bibinfo{person}{Marinka Zitnik}.} \bibinfo{year}{2020}\natexlab{}.
\newblock \showarticletitle{GNNGuard: Defending Graph Neural Networks against Adversarial Attacks}. In \bibinfo{booktitle}{\emph{Advances in Neural Information Processing Systems}}. \bibinfo{publisher}{{}}, \bibinfo{address}{{}}, \bibinfo{pages}{{}}.
\newblock


\bibitem[Zhu et~al\mbox{.}(2019)]%
        {kdd19-rgcn}
\bibfield{author}{\bibinfo{person}{Dingyuan Zhu}, \bibinfo{person}{Ziwei Zhang}, \bibinfo{person}{Peng Cui}, {and} \bibinfo{person}{Wenwu Zhu}.} \bibinfo{year}{2019}\natexlab{}.
\newblock \showarticletitle{Robust Graph Convolutional Networks Against Adversarial Attacks}.
\newblock \bibinfo{journal}{\emph{{}}}  \bibinfo{volume}{{}} (\bibinfo{year}{2019}), \bibinfo{pages}{{}}.
\newblock


\bibitem[Zhu et~al\mbox{.}(2020)]%
        {grace}
\bibfield{author}{\bibinfo{person}{Yanqiao Zhu}, \bibinfo{person}{Yichen Xu}, \bibinfo{person}{Feng Yu}, \bibinfo{person}{Qiang Liu}, \bibinfo{person}{Shu Wu}, {and} \bibinfo{person}{Liang Wang}.} \bibinfo{year}{2020}\natexlab{}.
\newblock \showarticletitle{Deep graph contrastive representation learning}.
\newblock \bibinfo{journal}{\emph{arXiv preprint arXiv:2006.04131}}  \bibinfo{volume}{{}} (\bibinfo{year}{2020}), \bibinfo{pages}{{}}.
\newblock


\bibitem[Zhu et~al\mbox{.}(2021)]%
        {www21-gca}
\bibfield{author}{\bibinfo{person}{Yanqiao Zhu}, \bibinfo{person}{Yichen Xu}, \bibinfo{person}{Feng Yu}, \bibinfo{person}{Qiang Liu}, \bibinfo{person}{Shu Wu}, {and} \bibinfo{person}{Liang Wang}.} \bibinfo{year}{2021}\natexlab{}.
\newblock \showarticletitle{Graph contrastive learning with adaptive augmentation}. In \bibinfo{booktitle}{\emph{Proceedings of the Web Conference 2021}}. \bibinfo{publisher}{{}}, \bibinfo{address}{{}}, \bibinfo{pages}{2069--2080}.
\newblock


\bibitem[Z{\"u}gner et~al\mbox{.}(2018)]%
        {18kdd-nettack}
\bibfield{author}{\bibinfo{person}{Daniel Z{\"u}gner}, \bibinfo{person}{Amir Akbarnejad}, {and} \bibinfo{person}{Stephan G{\"u}nnemann}.} \bibinfo{year}{2018}\natexlab{}.
\newblock \showarticletitle{Adversarial attacks on neural networks for graph data}. In \bibinfo{booktitle}{\emph{Proceedings of the 24th ACM SIGKDD International Conference on Knowledge Discovery \& Data Mining}}. \bibinfo{publisher}{{}}, \bibinfo{address}{{}}, \bibinfo{pages}{2847--2856}.
\newblock


\bibitem[Z{\"u}gner and G{\"u}nnemann(2019)]%
        {iclr18-mettack}
\bibfield{author}{\bibinfo{person}{Daniel Z{\"u}gner} {and} \bibinfo{person}{Stephan G{\"u}nnemann}.} \bibinfo{year}{2019}\natexlab{}.
\newblock \showarticletitle{Adversarial attacks on graph neural networks via meta learning}.
\newblock \bibinfo{journal}{\emph{arXiv preprint arXiv:1902.08412}}  \bibinfo{volume}{{}} (\bibinfo{year}{2019}), \bibinfo{pages}{{}}.
\newblock


\end{thebibliography}

\end{document}